\documentclass[a4paper,fleqn]{cas-dc}
\usepackage{amssymb}
\usepackage{multirow} 
\usepackage{graphicx} 
\usepackage{xcolor} 
\usepackage{hyperref}
\usepackage{amssymb}
\usepackage{amsthm, amsmath}
\usepackage{mathrsfs}
\usepackage{graphicx}
\usepackage{subfig}
\usepackage{natbib}

\usepackage{algorithm}
\usepackage{algpseudocode} 

\def\tsc#1{\csdef{#1}{\textsc{\lowercase{#1}}\xspace}}
\tsc{WGM}
\tsc{QE}
\tsc{EP}
\tsc{PMS}
\tsc{BEC}
\tsc{DE}

\begin{document}
\let\WriteBookmarks\relax
\def\floatpagepagefraction{1}
\def\textpagefraction{.001}

\title [mode = title]{
Utilizing Contextual Clues and Role Correlations for Enhancing Document-level Event Argument Extraction
}

\shorttitle{}

\shortauthors{}



\author[1]{Wanlong Liu}[style=chinese]
\credit{Conceptualization, Methodology, Experimentation, Writing - Original draft preparation}

\author[1]{Dingyi Zeng}[style=chinese]
\credit{Conceptualization, Methodology, Experimentation, Writing - Original draft preparation}

\author[1]{Li Zhou}[style=chinese]
\credit{Conceptualization, Experimentation, Writing - Original draft preparation}

\author[1]{Yichen Xiao}[style=chinese]
\credit{Methodology, Investigation}

\author[1]{Shaohuan Cheng}[style=chinese]
\credit{Conceptualization, Supervision}

\author[1]{Weishan Kong}[style=chinese]
\credit{Conceptualization, Supervision}


\author[1]{Hongyang Zhao}[style=chinese]
\credit{Conceptualization, Supervision}

\author[1]{Malu Zhang}[style=chinese]
\credit{Supervision, Investigation}

\author[1]{Wenyu Chen}[style=chinese, orcid=https://orcid.org/0000-0002-9933-8014]
\ead{cwy@uestc.edu.cn}
\cormark[1]
\credit{Conceptualization, Supervision, Funding acquisition}

\address[1]{Department of Computer Science and Engineering, University of Electronic Science and Technology of China, Chengdu, 611731, China}


\begin{abstract}
Document-level event argument extraction is a crucial yet challenging task within the field of information extraction. Current mainstream approaches primarily focus on the information interaction between event triggers and their arguments, facing two limitations: 
insufficient context interaction and the ignorance of event correlations. Here, we introduce a novel framework named CARLG (\textbf{C}ontextual \textbf{A}ggregation of Clues and \textbf{R}ole-based \textbf{L}atent \textbf{G}uidance), comprising two innovative components: the Contextual Clues Aggregation (CCA) and the Role-based Latent Information Guidance (RLIG). The CCA module leverages the attention weights derived from a pre-trained encoder to adaptively assimilate broader contextual information, while the RLIG module aims to capture the semantic correlations among event roles. We then instantiate the CARLG framework into two variants based on two types of mainstream EAE approaches. Notably, our CARLG framework introduces less than 1\% new parameters yet significantly improving the performance. Comprehensive experiments across the RAMS, WikiEvents, and MLEE datasets confirm the superiority of CARLG, showing significant superiority in terms of both performance and inference speed compared to major benchmarks.  Further analyses demonstrate the effectiveness of the proposed modules.

\end{abstract}


\maketitle

\section{Introduction}
\label{sec:intro} 

\begin{figure}[tbp]
    \centering
    \includegraphics[width=1.0\linewidth]{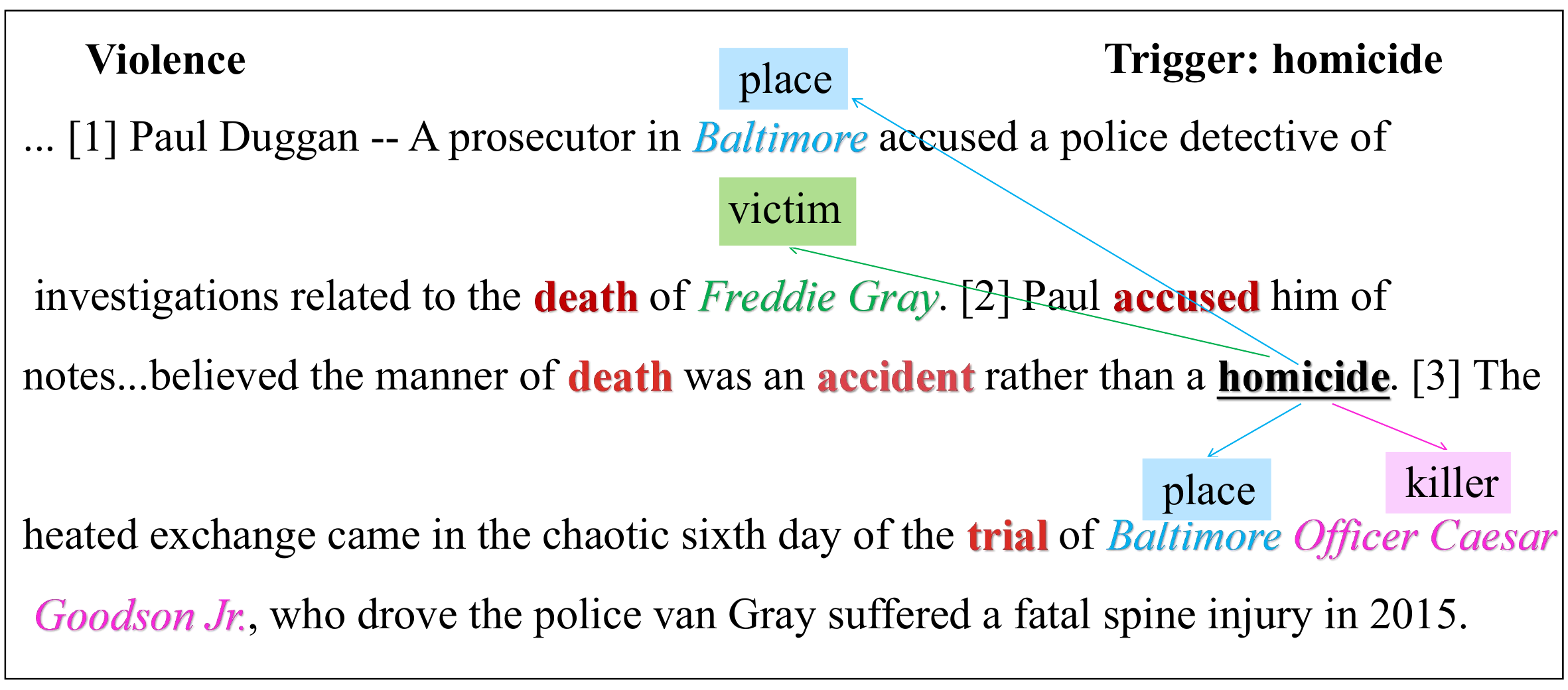}
    
\caption{A document from RAMS~\citep{ebner2020multi} dataset.
The event \texttt{violence} has the trigger of \textit{Homicide} and there are three distinct arguments, each assuming different roles. 
We use the color red to emphasize specific words, representing contextual clues that are significant for the extraction of arguments. 
}
    \label{fig:fig1}
\end{figure}

Within the domain of Information Extraction, document-level event argument extraction (EAE) is a vital task focused on identifying event arguments and their corresponding roles within textual content associated with specific events. This task is a key process of event extraction~\citep{xu2023multi, lu2022explainable, lv2022trigger} and has broad applications across various fields. For instance, it is instrumental in the analysis of social media networks~\citep{mitchell1974social, DING2023110687, YANG2022109939}, enhancing recommendation systems ~\citep{lietal2020gaia, LI2024111124, DONG2022108954}, and improving dialogue systems ~\citep{Zhang2020.08.03.20167569, ZHAO2023110927}. As illustrated in Fig.~\ref{fig:fig1}, with the event trigger ``homicide'' of event type \texttt{Violence}, the goal of EAE system is to identify argument ``Freddie Gray'' as the role \textit{victim}, ``Baltimore'' as the \textit{place}, and ``Officer Caesar Goodson Jr.'' as the \textit{killer}.

Compared with sentence-level EAE~\citep{liuetal2018jointly, waddenetal2019entity, tongetal2020improving}, document-level EAE faces more complex issues, notably long-range dependencies and the need for inferences across multiple sentences, as discussed in research~\citep{ebner2020multi} and~\citep{lietal2021document}. To solve these problems, some prompt-based methods~\citep{ma2022prompt, hsu2023ampere} design suitable templates for long text, which solves the long-distance dependency problem and extract arguments in a generative slot-filling manner. Additionally, some works utilize graph structures employing heuristic approaches~\citep{zhang2020two, pouranbenveysehetal-2022-document}, syntactical frameworks~\citep{xuetal2021document, liu2022document, 9204800}, and  Graph Neural Networks~\citep{kipfsemi} for the facilitation of reasoning across sentences.  However, current mainstream approaches ignore two essential aspects: (a) the contextual clue information; (b) the semantic correlations among roles. 

Contextual clues refer to text elements in the document, excluding the target arguments themselves, which provide beneficial guidance for predicting complex argument roles. These clues play a significant role in event argument extraction by offering valuable information. For instance, in Figure~\ref{fig:fig1}, the contextual clues, such as \textit{death}, \textit{accused}, and \textit{trial}, serve as important indicators for identifying the arguments ``Freddie Gray'' and ``Officer Caesar Goodson Jr'' within the event \texttt{Violence}. By considering these contextual clues, the model can leverage wider context information to make more informed and accurate predictions.
However, many prior studies~\citep{xu2022two, lietal2021document, liu-etal-2023-enhancing-document} just rely on the pre-trained transformer encoder to  capture global context information in text implicitly, ignoring the necessity of  context clues information that is highly relevant to the entity~\citep{zhou2021document} and target event~\citep{ebner2020multi}.
Therefore,
in this paper, we propose a \textbf{C}ontextual \textbf{C}lues
\textbf{A}ggregation
(CCA) module, which leverages the contextual attention product from the  pre-trained transformer encoder to aggregate the event-specific  contextual clues for each argument, enhancing the candidate argument representation by incorporating broader and relevant contextual information. 

\begin{figure}[tbp]
    \centering
    \includegraphics[width=1.0\linewidth]{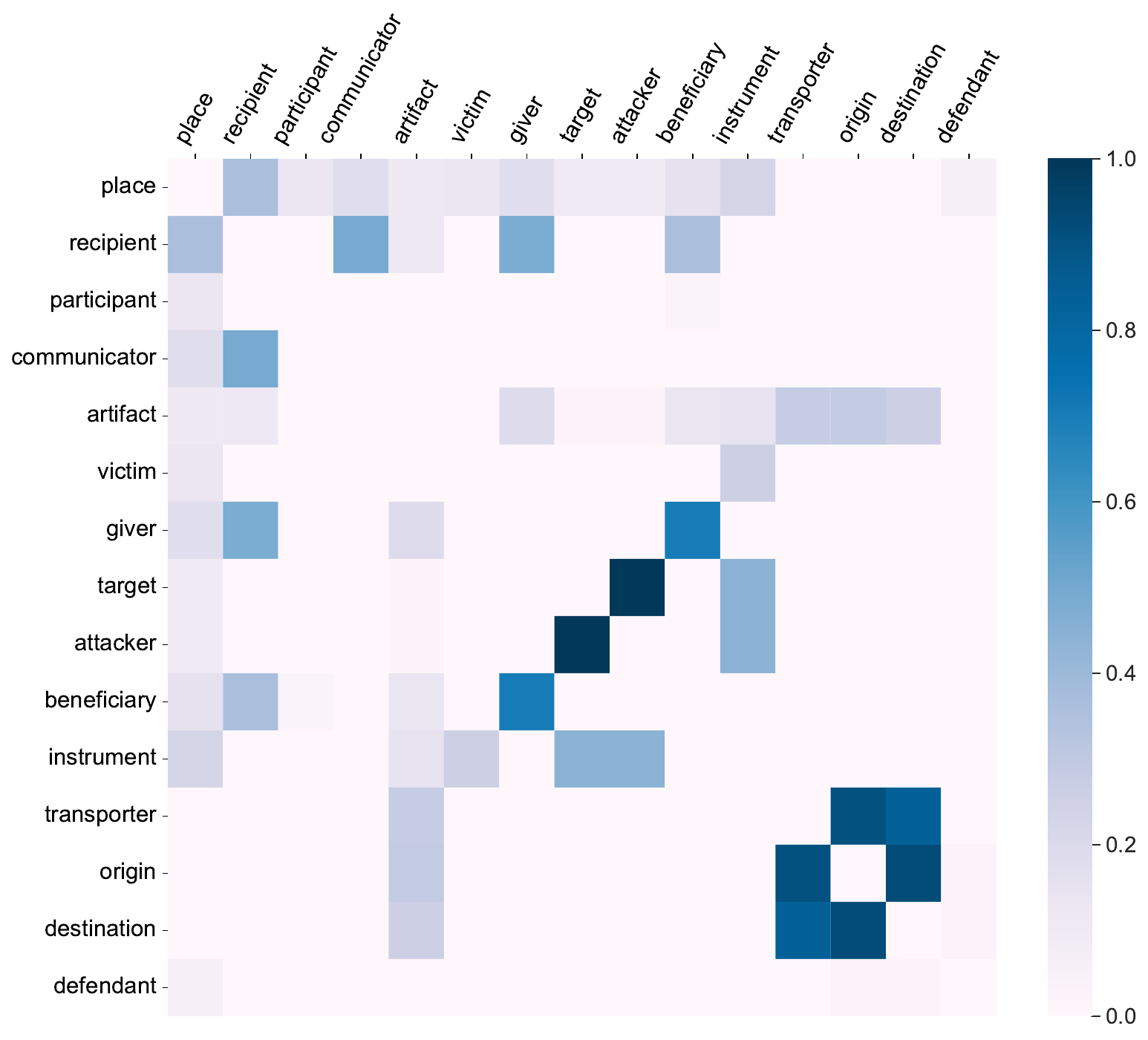}
    
    \caption{The visualization represents the co-occurrence frequencies of the top 15 common roles within the RAMS test set, which is defined as the ratio between the co-occurrence times and the sum of the total occurrence times of two roles, and the self-co-occurrence count is set to zero.
    }
    \label{fig:fig2}
\end{figure}

For real-world document-level EAE task, numerous event roles exhibit semantic correlations.
For example, as illustrated in Figure~\ref{fig:fig1}, there exists a strong semantic correlation between the roles of \textit{killer} and \textit{victim}, serving as beneficial guidance for extracting arguments associated with these roles within the target event \texttt{Violence}. In addition, many roles co-occur in multiple events, indicating strong potential for close semantic correlations among them.
For a more in-depth analysis, we perform a frequency count and visualize the co-occurrence of the top 15 roles in the RAMS dataset, as shown in Figure~\ref{fig:fig2}.
Notably, roles such as \textit{attacker}, \textit{target} and \textit{instrument} show high co-occurrence frequency, suggesting their strong semantic correlations compared with other roles.
Therefore, in this paper, we propose a \textbf{R}ole-based \textbf{L}atent \textbf{I}nformation \textbf{G}uidance (RLIG) module, comprising role-interactive encoding, the fusion of role information and latent role guidance.  Specifically, in role-interactive encoding, we add roles into the input sequence, enabling role embeddings to learn both semantic information and latent correlations among roles through self-attention mechanism. Subsequently, the latent role embeddings are combined with candidate argument spans via pooling and concatenation operations. Finally, we achieve alignment of argument spans, triggers, and roles within a unified space through a novel alignment mechanism utilizing Tucker Decomposition~\citep{tucker1964extension}, which provides valuable information guidance for document-level EAE.

In summary, in this paper, we propose the CARLG framework (\textbf{C}ontextual Clues
\textbf{A}ggregation and \textbf{R}ole-based \textbf{L}atent Information \textbf{G}uidance), consisting of CCA module and RLIG module. Each module leverages the fine-organized attentive weights obtained from the transformer-based pre-trained  model, introducing less than 1\% of new parameters. We instantiate the CARLG framework into two variants: the span-based $\text{CARLG}_\text{span}$ and the prompt-based $\text{CARLG}_\text{prompt}$, based on two types of current mainstream EAE approaches. 
Both variants demonstrate significant superiority over leading benchmarks, in terms of both performance and inference speed. 
Our contributions are summarized as follows:

\begin{itemize} 
\item To capture the contextual clues within the text, we propose a Contextual Clues Aggregation (CCA) module. It adaptively identifies and aggregates the contextual clue information and enhances the representation of candidate arguments, achieving sufficient context interaction for document-level EAE.

\item We  propose a Role-based Latent Information Guidance (RLIG) module to capture event role correlations, which employs learnable latent role embeddings to enhance the semantic interaction among event roles.

\item  We instantiate the CARLG framework into two variants: the $\text{CARLG}_\text{span}$ and the $\text{CARLG}_\text{prompt}$, based on two types of current mainstream EAE approaches, which shows the adaptability and transferability of our framework.

\item  Extensive experiments on the public RAMS and WikiEvents~\citep{lietal2021document} and MLEE~\citep{10.1093/bioinformatics/bts407} datasets demonstrate that both CARLG variants show significant superiority in both performance and efficiency.  Comprehensive ablation studies and analyses illustrate the effectiveness of our proposed framework.
\end{itemize} 

\section{Related Works}

In this section, we provide a brief overview of the related works on the EAE task. We analyze the strengths and weaknesses of current mainstream EAE methods and emphasize the importance of considering contextual clues and role embedding in this task. Additionally, we discuss several applications in the NLP field that leverage contextual clues and relation embedding, which motivates our proposed approach.

\subsection{Event Argument Extraction}

Event argument extraction (EAE) is pivotal for improving  event semantic understanding by identifying event arguments within a context, which constitutes a fundamental step in various downstream applications, including question answering~\citep{liu2022document, 6544211, 9257099}, text summarization~\citep{allahyari2017text, 8949697}, and assisted decision-making~\citep{zhang2020effect, 9339934}. 

Previous approaches primarily concentrate on  identifying the event trigger and its associated arguments from individual sentences. In \citep{chenetal2015event}, a neural pipeline model was first proposed for event extraction, which was later extended by \citep{ nguyengrishman2015event, liuetal2017exploiting} to incorporate recurrent neural networks (RNNs) and convolutional neural networks (CNNs) into the pipeline model. ~\citep{liuetal2018jointly, yanetal2019event} utilize dependency trees to construct semantic and syntactic dependencies and relations, which captures the dependency of words in a sentence. \citep{waddenetal2019entity} explores all possible spans and utilizes graph neural networks to construct span graphs for information propagation. Additionally, transformer-based pre-trained models ~\citep{waddenetal2019entity, tongetal2020improving, liuetal2022dynamic} achieve impressive performance in event extraction.

However, in real-world situations, elements of an event are often expressed across multiple sentences, leading to the emergence of document-level event extraction (DEE)~\citep{6910247, jiang2023hard, SU2024111180, SCABORO2023110675}. DEE seeks to identify event arguments throughout whole documents, addressing the issue of extended dependencies\citep{wangetal2022query, xu2022two}.
    We summarize these recent DEE works into 3 categories: (1)  Tagging-based methods. Works like \citep{wangetal2021cleve, ducardie2020document} apply the BiLSTM-CRF framework for document-level event extraction. Zheng~\citep{zhengetal2019doc2edag} introduces a mixed graph and a tracking module to capture the interrelations between events.\citep{lietal2021document} approaches the issue through conditional generation, whereas \citep{duetal2021grit} views it as a sequence-to-sequence endeavor. Additionally, Wei~\citep{weietal2021trigger} redefines the task in terms of a reading comprehension challenge. (2) Span-based methods. \citep{yangetal2021document} proposes an encoder-decoder framework for parallel extraction of structured events.  Works~\citep{ebner2020multi, zhang2020two} predict  argument roles for selected text segments under a maximum length restriction. Furthermore, Xu~\citep{xu2022two} suggests a dual-stream encoder complemented by an AMR-guided graph to manage extended dependencies. (3) Prompt-based methods.  
Ren~\citep{renetal2022clio} integrates argument roles into document encoding to account for multiple role details for nested arguments. \citep{ma2022prompt, liu2022document, XU2023110375} propose a prompt-based method that utilizes slotted prompts for generative slot-filling to extract arguments. Among them, 
span-based and prompt-based methods are current mainstream methods, which have been demonstrated superior generalizability and competitive
performance~\citep{hsu2023ampere}.  

Most existing document-level EAE methods  focus on the interaction between arguments and event triggers, ignoring the correlations of  roles among events, or the clue information provided by non-arguments context, which can be crucial for the task. Therefore, we attempt to enhance the document-level EAE model with useful contextual information and semantic role correlations.

\subsection{Contextual Clues}
Contextual clues refer to the information and cues present in the surrounding context of a given entity, event, or situation~\citep{jiang2022supervised, zhang2020contextual, liu2022document, li2024feature}. These clues provide additional context and can help in understanding the meaning, significance, or relationships associated with the target entity or event~\citep{peters2018deep}.  NC-DRE~\citep{zhang2022nc} introduces an encoder-decoder structure, leveraging a pre-trained encoding model and a graph neural network (GNN) for decoding. This setup facilitates the use of non-entity clues in document-level relation extraction. Concurrently, a method for localized context pooling~\citep{zhou2021document}  has been put forward, allowing for the direct application of attention mechanisms from pre-trained language models. This method efficiently leverages the  contextual information tied to particular entity pairs, aiding in capturing the beneficial context clues for determining relations. Xu~\citep{xu2021discriminative} uses the self-attention mechanism~\citep{vaswani2017attention} to learn a document-level context representation which contributes to the improvement of information extraction.  

Therefore, contextual clues contain important semantic information, which can provide information guidance for deep mining of semantic associations between entities.
In this paper, we are the first to incorporate contextual clues techniques into the event argument extraction task and propose the Contextual Clues Aggregation module to explicitly capture contextual clues.

\subsection{Relation Embedding}
Many methods in the field of information extraction leverage relation embeddings to enhance the effectiveness of various tasks~\citep{9086164, 9694521, 9123572}. CRE~\citep{chen-badlani-2020-relation} presents a novel approach to relation extraction by utilizing sentence representations as relation embeddings, which are assessed through entity embeddings trained concurrently with the knowledge base.~\citep{xuetal2022emrel} constructs relation representation explicitly and interacts it with entities and context, which achieves excellent results in multi-triple relation extraction.  ~\citep{han2022documentlevel} employs relation embeddings as a medium and introduces two sub-tasks focused on co-occurrence prediction, aiming to enhance the retrieval of relational facts.

Relation embedding is typically utilized in relation extraction tasks. In the case of event argument extraction, we can consider event roles as relations connecting arguments and events. This viewpoint enables the utilization of relation embedding to proficiently capture the semantic correlations between arguments and events within EAE endeavors.
Therefore,  we aim to use contextual
clues and role correlations embedded in latent embedding to strengthen the document-level EAE performance.

\begin{figure*}[tbp]
    \centering
    \includegraphics[width=0.8\linewidth]{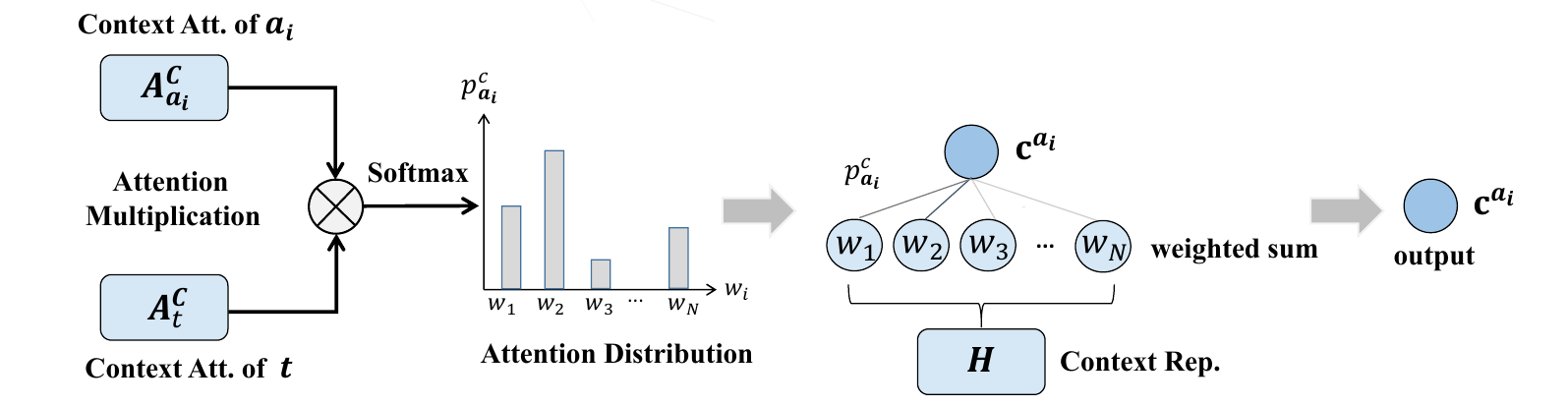}
    
    \caption{Contextual Clues Aggregation Module. 
The core process of the Role-based Latent Information Guidance algorithm is also similar to this.}
    \label{fig:fig4}
\end{figure*}
\section{Preliminaries}

\subsection{Important Notions}
We follow the definition of event argument extraction in RAMSv1.0 
benchmark~\citep{ebner2020multi}.
To better understand this task, we first clarify  some important notions.
\begin{itemize} 
\item \textbf{Event mention}:  a text span within a document that mentions an entity object,  e.g., the phrase \textit{Paul Duggan} in Figure~\ref{fig:fig1}.

\item \textbf{Event trigger (trigger)}:  the key term that most explicitly represents the manifestation of the event, e.g., the word \textit{homicide} in Figure~\ref{fig:fig1}.

\item \textbf{Event type}:  the semantic class of an event, e.g., the event type \textit{Violence} in Figure~\ref{fig:fig1}.

\item \textbf{Argument role (role)}:  the relation between the event and the  argument in which it is involved, for example, the role of \textit{Baltimore} is identified as \textit{Place} in Figure~\ref{fig:fig1}.
For this task, argument roles are predetermined for every event type.

\item \textbf{Event argument (argument)}: an entity mention, value or temporal expression playing a specific argument role, e.g., the argument \textit{Freddie Gray} in Figure~\ref{fig:fig1}.

\end{itemize}

\subsection{Task Formulation}

Formally, for a document $\mathcal{D}$ composed of $N$ words, denoted as $\mathcal{D} = \{w_1, w_2, \ldots, w_N\}$, a set of pre-defined event types $\mathcal{E}$, a corresponding set of argument roles $\mathcal{R}_e$ associated with each event $e \in \mathcal{E}$, and the event trigger $t$ in $\mathcal{D}$. The goal is to detect every pair $(r, s)$ for every event $e$ within the document $\mathcal{D}$, where $r \in \mathcal{R}_e$ signifies a role associated with the event $e$, and $s$ denotes a contiguous text span within $\mathcal{D}$. In alignment with earlier research~\citep{xu2022two, ebner2020multi}, we extract event arguments independently for each event within a document.

\section{CARLG}

In this section, we introduce CARLG framework, comprising Contextual Clues
Aggregation (CCA) module which adaptively aggregates contextual clues and Role-based Latent Information Guidance (RLIG) module which effectively captures semantic
correlations among roles.  
The core purpose of these two modules is to enhance the representation of the arguments to be predicted. Additionally, they can be applied for any approach that relies on argument representation, including span-based methods that make predictions using candidate span representation, and prompt-based methods that predict based on argument slot representation. 

\subsection{Contextual Clues Aggregation}

Given that different events focus on different contextual information, the CCA module adaptively aggregates contextual information based on specific event feature to obtain a broader context representation. Since the event trigger is crucial for the arguments within an event, we adaptively aggregate context information based on each trigger-argument pair, denoted as $(t, a_i)$ and $a_i$ is the $i^{th}$ argument.  We leverage the attentive heads from the pre-trained transformer encoder, which enables the adoption of attentions from the pre-trained language model instead of training new attention layers. Figure~\ref{fig:fig4} illustrates the process of the proposed CCA module.

Specifically, we encode the input text sequence $S$ and get the encoded representation $\textbf{H}^C$ through a transformer-based pre-trained language model. We can obtain the  attention head matrix $\textbf{A}^{C} \in \mathbb{R}^{H\times l\times l}$ from the final transformer layer, where $H$ is the number of attention heads and $l$ represents the text sequence length.
 We then extract the contextual attention heads of the trigger  and each argument through the pooling operation, denoted as $\textbf{A}^{C}_{t} \in \mathbb{R}^{l}$ and $\textbf{A}^{C}_{a_i} \in \mathbb{R}^{l}$.
It's important to note that since we do not know in advance what the argument is, the way it is obtained varies by method. For span-based methods, we consider the tokens within the span as the argument. For prompt-based methods, we treat the slot in the prompt template as the argument. We explain this in detail in Section~\ref{Adapting CARLG to Span-based and Prompt-based EAE Approaches}.

Then for each argument, we get the contextual clue information  $\textbf{c}_{a_i} \in \mathbb{R}^{d}$ by applying multiplication to the attentions and then normalizing the result:

\begin{equation}
\centering
\label{eq4}
    \begin{array}{c}
    \textbf{p}^{C}_{a_i}=softmax(\textbf{A}^{C}_{a_i}\cdot \textbf{A}_{t}^{C}\;)\vspace{1.3ex}, \\
    \textbf{c}_{a_i}={\mathbf{H}^C}^T\ \textbf{p}^{C}_{a_i},
    \end{array}
\end{equation}%
where $\cdot$ means dot product and $\textbf{p}^{C}_{a_i} \in \mathbb{R}^{l_w}$ is the computed contextual attention weight vector. $T$ is the transpose symbol.
$\textbf{c}_{a_i} $ is the enhanced context vector specific to the argument $a_i$ and trigger $t$. The detailed algorithm is shown in Algorithm~\ref{Algorithm 1}.

\begin{algorithm}
\caption{Contextual Clues   Aggregation}
\label{Algorithm 1}
\begin{algorithmic}[1]
\Require{Input text sequence $S$, pre-trained transformer encoder $Encoder$}
\Ensure{Context-enhanced argument representations $\textbf{c}_{a_i}$}
\State $\textbf{H}^C \gets Encoder(S)$ \Comment{Encode $S$ using the pre-trained transformer}
\State Obtain attention head matrix $\textbf{A}^{C} \in \mathbb{R}^{H\times l\times l}$ from the final layer of $Encoder$
\For{each argument $a_i$}
    \State Identify the index $t_{idx}$ of the trigger in the sequence
    \State Identify the index range $a_{i_{idx}}$ of the argument $a_i$ in the sequence
    \State Extract $\textbf{A}^{C}_{t} \in \mathbb{R}^{l}$ for the trigger by selecting the $t_{idx}$-th row from $\textbf{A}^{C}$
    \State Extract $\textbf{A}^{C}_{a_i} \in \mathbb{R}^{l}$ for the argument $a_i$ by aggregating rows in $a_{i_{idx}}$ from $\textbf{A}^{C}$
    \State Compute contextual attention weights: $\textbf{p}^{C}_{a_i} = \text{softmax}(\textbf{A}^{C}_{a_i} \cdot \textbf{A}_{t}^{C})$
    \State Obtain context vector: $\textbf{c}_{a_i} = {\mathbf{H}^C}^T \textbf{p}^{C}_{a_i}$
\EndFor
\State \Return {$\textbf{c}_{a_i}$ for each argument $a_i$}
\end{algorithmic}
\end{algorithm}

\subsection{Role-based Latent Information Guidance}

In this section, we propose Role-based Latent Information Guidance (RLIG) module, consisting of Role-interactive Encoding and Role Information Fusion.
\subsubsection{Role-interactive Encoding}

In NLP applications, the inclusion of prefixes, like relation and other injected information, significantly enhances the performance of downstream tasks~\citep{xuetal2022emrel, hsu2023ampere, liu-etal-2023-document}. To effectively capture the semantic correlations between roles, we integrate the information of role types as prefixes into the input and utilize multi-head attention to facilitate interaction between the context and roles.
  We generate latent role embeddings by assigning unique special tokens ~\footnote{In our implementation, using \texttt{[unused]} tokens in BERT~\citep{devlinetal2019bert}  and introducing specific tokens for RoBERTa~\citep{DBLP:journals/corr/abs-1907-11692}} to represent different roles in the framework of a pre-trained model, with each role type gaining a unique latent representation. Considering that role names carry valuable semantic information~\citep{wangetal2022query}, we encase role names in special tokens indicative of their role types, selecting the embedding of the initial special token as the role's representation. This approach ensures the capture of unique attributes and subtle semantic distinctions of each role. For instance, the role  \textit{``Victim''} is denoted as \text{``$[R_0] \quad Victim \quad [R_0]$''}, with \text{``$[R_0]$''} representing the specialized token for \textit{``Victim''}.

\begin{algorithm}
\caption{Role-based Latent Information Guidance}
\label{Algorithm 2}
\begin{algorithmic}[1]
\Require{Input text sequence $S$, set of roles $\{r_0, r_1, \ldots, r_n\}$, pre-trained transformer encoder}
\Ensure{Role-enhanced argument representations $\textbf{r}_{a_i}$}
\State Enhance $S$ by concatenating role type elements: \\
$S' = [R_0] \ {r_0} \  [R_0] \  [R_1] \  {r_1} \  [R_1] \  ... \  \mathrm{[SEP]} \  w_1 \ w_2 \ ... \  w_N$
\State Encode $S'$ using the pre-trained transformer to obtain $\mathbf{H}^C$ and $\mathbf{H}^R$
\For{each argument $a_i$}
    \State Identify indices $t_{idx}$ and $a_{i_{idx}}$ for trigger $t$ and argument $a_i$ in $S'$
    \State Extract $\textbf{A}^{R}_{t} \in \mathbb{R}^{l_r}$ by selecting the attention vector corresponding to \\
    \quad \quad $t_{idx}$ from $\mathbf{H}^R$
    \State Extract $\textbf{A}^{R}_{a_i} \in \mathbb{R}^{l_r}$ by selecting the attention vector corresponding to \\
    \quad \quad $a_{i_{idx}}$ from $\mathbf{H}^R$
    \State Compute role attention weights: $\textbf{p}^{R}_{a_i} = \text{softmax}(\textbf{A}^{R}_{a_i} \cdot \textbf{A}^{R}_{t})$
    \State Obtain role-specific information: $\textbf{r}_{a_i} = {\mathbf{H}^R}^T \textbf{p}^{R}_{a_i}$
\EndFor
\State \Return {$\textbf{r}_{a_i}$ for each argument $a_i$}
\end{algorithmic}
\end{algorithm}

Given the input sequence $S$, we enhance it by concatenating role type  elements in the following manner:
\[\quad   S \ =  \ \textsc{ $[R_0] \ {r_0} \   [R_0] \  [R_1] \  {r_1} \  [R_1] \  ... \  \mathrm{[SEP]}$}  \] 
\[\ \quad \quad \quad \quad w_1 \ w_2 \ * \ t_e \ * \ ... \  w_N, \  \]

\noindent where \textit{$[R_0]$} and \textit{$[R_1]$} denote the special role type tokens for $r_0$ and $r_1$, respectively, and we utilize the final $\mathrm{[SEP]}$ token to signify the ``none'' category. By employing  role-interactive encoding, we are able to get the context representation $\mathbf{H}^C \in \mathbb{R}^{l_w\times d}$, and the role representation $\mathbf{H}^R \in \mathbb{R}^{l_r\times d}$. Here, $l_w$ denotes the length of the list of word pieces, and $l_r$ represents the length of the list of roles.

\subsubsection{Role Information Fusion}
Through role-interactive encoding, the role embeddings are able to capture semantic correlations and dynamically adapt to the target event and context. To ensure each candidate argument benefits from the guidance of role-specific information, we adapt our method for aggregating contextual clues to selectively incorporate role information. The aggregated role information $\textbf{r}_{a_i} \in \mathbb{R}^{d}$ for argument ${a_i}$ is obtained as follows:

\begin{equation}
\label{eq71}
\setlength\abovedisplayskip{3pt plus 3pt minus 7pt}
\setlength\belowdisplayskip{3pt plus 3pt minus 7pt}
    \begin{array}{c}

\textbf{p}^{R}_{a_i}=softmax(\textbf{A}^{R}_{a_i}\cdot \textbf{A}_{t}^{R}\;)\vspace{1.3ex}, \\
    \textbf{r}_{a_i}={\mathbf{H}^R}^T \ \textbf{p}^{R}_{a_i},
    \end{array}
\end{equation}%
where $\textbf{A}^{R}_{a_i} \in \mathbb{R}^{l_r}$ represents the role attention for  argument $a_i$, while $\textbf{A}^{R}_{t} \in \mathbb{R}^{l_r}$ denotes the role attention associated with the trigger $t$. Additionally, the vector $\textbf{p}^{r}_{a_i} \in \mathbb{R}^{l_r}$ characterizes the computed attention weight vector for roles.
$\textbf{r}_{a_i} $ is the aggregated role vector for argument $a_i$. The detailed algorithm is shown in Algorithm~\ref{Algorithm 2}.

Then the enhanced context
vector $\textbf{c}_{a_i}$ and role vector $\textbf{r}_{a_i}$ can be utilized to enhance the representation e of the argument $a_i$ to be predicted.

\section{Adapting CARLG to Span-based and Prompt-based EAE Approaches}
\label{Adapting CARLG to Span-based and Prompt-based EAE Approaches}
We instantiate the CARLG framework into two variants: the span-based $\text{CARLG}_\text{span}$ and the prompt-based $\text{CARLG}_\text{prompt}$, based on two types of current mainstream EAE approaches.

\subsection{$\text{CARLG}_\text{span}$}
\begin{figure*}[tbp]
    \centering
    \includegraphics[width=0.9\linewidth]{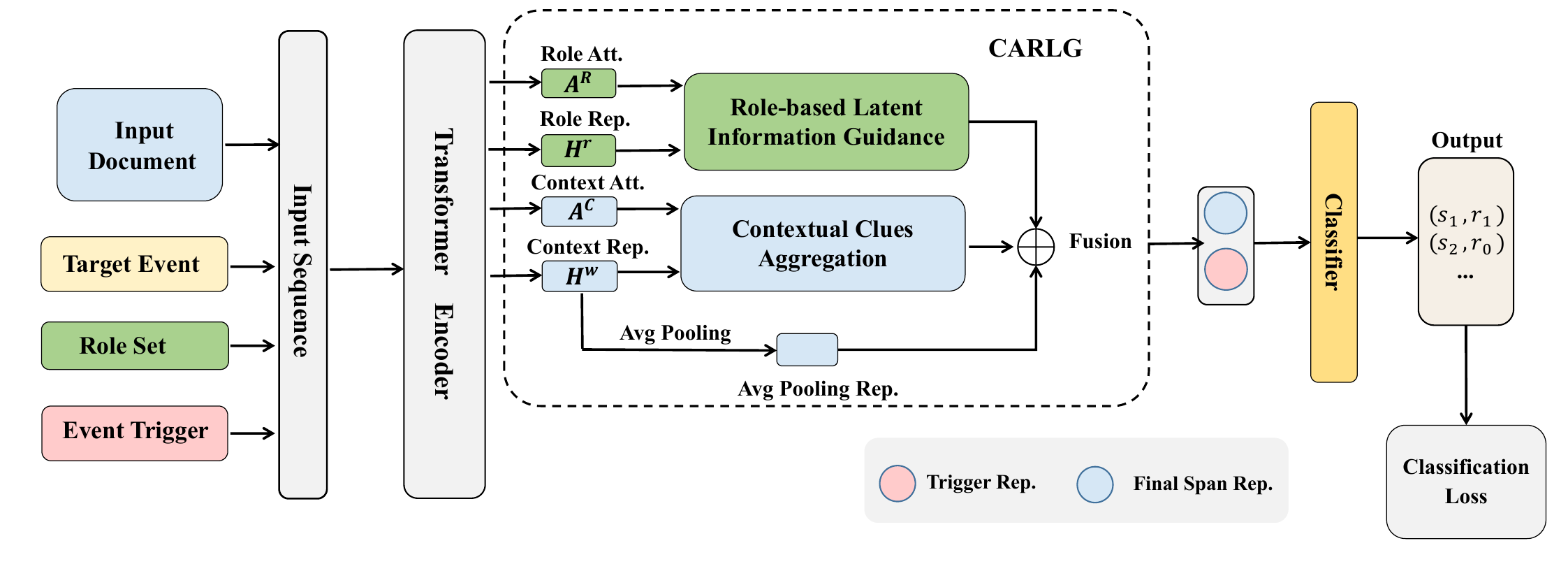}
    
    \caption{The main structure of $\text{CARLG}_\text{span}$. The input sequence, encompassing roles and event types, passes through a transformer-based encoder. This encoder generates output in the form of context representations, role representations, and attention heads. The Contextual Clue Aggregation (CCA) module dynamically consolidates contextual clues, while the Role-based Latent Information Generation (RLIG) module generates latent role embeddings and captures role correlations. Subsequently, the context vector is combined with the latent role vector to create the ultimate span representation. Finally, the classification module leverages this representation to forecast argument roles for all potential spans.}
    \label{fig:fig3}
\end{figure*}

We apply our CARLG framework to span-based EAE approaches, where the architecture is shown in Figure~\ref{fig:fig3}. 
 Since the application of CARLG framework is consistent across span-based methods, we choose the architecture of TSAR~\citep{xu2022two} as the foundation to illustrate the span-based baseline\footnote{We extract the core architecture of TASR as the span-based baseline, removing the two-stream module and the AMR graph module.}, and we detail the application of our CARLG framework to other typical span-based methods in the experimental Section~\ref{exp:main}.

\subsubsection{Role-interactive Encoder}
\label{sec:encoder}
We apply Algorithm~\ref{Algorithm 2} and construct a role-interactive encoder to capture role correlations. Following~\citep{xu2022two, ebner2020multi}, we divide a document into several instances, where each instance only contains one event. For an instance,
given the input document $\mathcal{D}$, the target event $e$ and its associated set of roles $\mathcal{R}_e = \{{r}_1, {r}_2, {r}_3, ...\}$, we enhance the input by concatenating these elements in the following manner:
\[\quad   S_\text{span} \ =  \ \mathrm{[CLS]} \  [E_e] \ {e}  \ [E_e] \ \mathrm{[SEP]} \  w_1 \ w_2 \ * \ t_e \ * \ ... \  w_N \  \]
\[ \quad \quad \mathrm{[SEP]}\textsc{  \ $[R_0] \ {r_0} \   [R_0] \  [R_1] \  {r_1} \  [R_1] \  ... \  \mathrm{[SEP]}$}, \] 

\noindent where \textit{$[R_0]$} and \textit{$[R_1]$} denote the special role type tokens for $r_0$ and $r_1$, respectively, and we utilize the final $\mathrm{[SEP]}$ token to signify the ``none'' category.
Next, we use a pre-trained transformer encoder to get the embedding of each token as follows:

\begin{equation}
    \mathbf{H} = \mathrm{Encoder} (S_\text{span}).
\end{equation}

We can get the event representation $\mathbf{H}^e \in \mathbb{R}^{1\times d}$ for the start \textsc{$[E_e]$} and the context representation $\mathbf{H}^w \in \mathbb{R}^{l_w\times d}$, where $l_w$ represents the length of the word pieces list. Notably, when the  input sequence is longer than the maximum token length, we employ a dynamic window approach~\citep{zhou2021document, zhang2021document} to process the long sequence. We divide the sequence into overlapping windows and calculate the embeddings for each window. Next, we obtain the final representation by averaging the embeddings of tokens that overlap across various windows.

\subsubsection{Contextual Clues Aggregation}

We apply Algorithm~\ref{Algorithm 1} to the span-based representations, capturing context clues to enhance the representation of candidate spans.
We employ the token-level attention heads $\textbf{A}^{C} \in \mathbb{R}^{H\times l_w\times l_w}$ obtained from the final transformer layer of the pre-trained language model. Here, $H$ represents the number of attention heads, and $l_w$ signifies the length of word pieces. Utilizing these attention heads, we compute the context attention $\textbf{A}^{C}_{i:j} \in \mathbb{R}^{l_w}$ for each candidate span spanning from $w_i$ to $w_j$, by applying average pooling:

\begin{equation}
\label{eq3}
     \textbf{A}^{C}_{i:j} = \frac1{H(j-i+1)}\sum_{h=1}^H\sum_{m=i}^j\mathbf{A}^{C}_{h,m}.
\end{equation}

Then for span-trigger tuple $(s_{i:j}, t)$, we get the contextual clue information  $\textbf{c}_{s_{i:j}} \in \mathbb{R}^{d}$ which is crucial for the candidate span, by applying multiplication to the attentions and then normalizing the result:

    

\begin{equation}
\setlength\abovedisplayskip{3pt plus 3pt minus 7pt}
\setlength\belowdisplayskip{3pt plus 3pt minus 7pt}
    \begin{array}{c}
    \textbf{p}^{C}_{i:j}=softmax(\textbf{A}^{C}_{i:j}\cdot \textbf{A}_{t}^{C}\;)\vspace{1.3ex}, \\
    \textbf{c}_{s_{i:j}}={\mathbf{H}^C}^T\ \textbf{p}^{C}_{i:j},
    \end{array}
\end{equation}%
where $\textbf{A}^{C}_{t} \in \mathbb{R}^{l_w}$ is the contextual attention of trigger $t$, $\cdot$ means dot product and $\textbf{p}^{C}_{i:j} \in \mathbb{R}^{l_w}$ is the computed contextual attention weight vector. $T$ is the transpose symbol.

\begin{figure*}[tbp]
    \centering
    \includegraphics[width=0.95\linewidth]{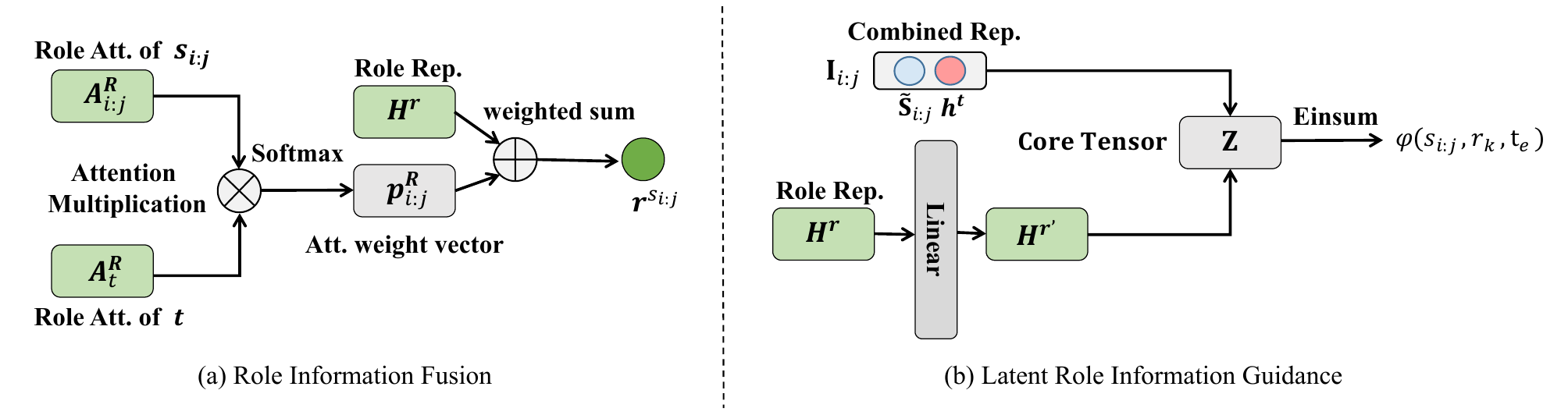}
    
    \caption{Detailed RLIG module for $\text{CARLG}_\text{span}$. Sub-figure (a) illustrates the information fusion and (b) illustrates the Latent Role Information Guidance operation.}
    \label{fig:fig5}
\end{figure*}

\subsubsection{Latent Role Information Guidance}
Since span-based methods ultimately classify candidate spans based on the  categories of roles, we propose Role Information Guidance on top of Algorithm~\ref{Algorithm 2} to better utilize the learned latent role representations, as shown in Figure~\ref{fig:fig5}.

\textbf{Latent Role Information Fusion} \quad
Through the role-interactive encoding, the role embeddings are able to capture semantic correlations and dynamically adapt to the target event and context. To ensure each candidate argument benefits from the guidance of role-specific information, we adapt our method for aggregating contextual clues to selectively incorporate role information. The aggregated role information $\textbf{r}_{s_{i:j}} \in \mathbb{R}^{d}$ for ${s_{i:j}}$ is obtained through contextual pooling, which involves adjusting the procedure described in Algorithm~\ref{Algorithm 2}:

\begin{equation}
\label{eq7}
\setlength\abovedisplayskip{3pt plus 3pt minus 7pt}
\setlength\belowdisplayskip{3pt plus 3pt minus 7pt}
    \begin{array}{c}

    \textbf{A}^{R}_{i:j} = \frac1{H(j-i+1)}\sum_{h=1}^H\sum_{m=i}^j\mathbf{A}^{R}_{h,m}\vspace{1.3ex},\\
    \textbf{p}^{R}_{i:j}=softmax(\textbf{A}^{R}_{i:j}\cdot \textbf{A}_{t}^{R}\;)\vspace{1.3ex}, \\
    \textbf{r}_{s_{i:j}}={\mathbf{H}^R}^T \ \textbf{p}^{r}_{i:j},
    \end{array}
\end{equation}%
where $\textbf{A}^{R} \in \mathbb{R}^{H\times l_w\times l_r}$ represents the role attention heads obtained from the final transformer layer of the pre-trained model. $\textbf{A}^{R}_{i:j} \in \mathbb{R}^{l_r}$ represents the role attention for each candidate span, while $\textbf{A}^{R}_{t} \in \mathbb{R}^{l_r}$ denotes the role attention associated with the trigger $t$. Additionally, the vector $\textbf{p}^{R}_{i:j} \in \mathbb{R}^{l_r}$ characterizes the computed attention weight vector for roles.
We can fuse the contextual clue information  $\textbf{c}_{s_{i:j}}$ and aggregated role information $\textbf{r}_{s_{i:j}}$into the average pooling representation as follows:

\begin{equation} 
\label{eq13}
\textbf{s}_{i:j}=tanh(\mathbf{W}^{Span}[\frac1{j-i+1}\sum_{k=i}^j\mathbf{h}_k^w; \textbf{c}_{s_{i:j}}; \textbf{r}_{s_{i:j}}]),
\end{equation}
where $\mathbf{W}^{Span} \in \mathbb{R}^{3d \times d}$ is learnable parameter. $\textbf{s}_{i:j}$ is the enhanced representation of the candidate span $s_{i:j}$.

\textbf{Role Information Guidance} \quad To maximize the utilization of the learned role representations, we introduce the Latent Role Information Guidance algorithm, tailored to the span-based architecture.
We first map the original $d$-dimensional role features into a reduced dimension $d'$ via a fully connected network to obtain the latent role information $\mathbf{H}^{r'} \in \mathbb{R}^{l_r\times d^{'}}$ as follows:

\begin{equation}
\label{eq5}
     \mathbf{H}^{r'} = \mathbf{H}^r  \mathbf{W}_3 + \mathbf{b}_w,
\end{equation}
where $\mathbf{W}_3 \in \mathbb{R}^{d\times d^{'}}$ and $\mathbf{b}_w \in \mathbb{R}^{d^{'}}$ are learnable parameters.
To fully utilize the expressive power of role embeddings, we propose a factorization-based alignment approach using Tucker decomposition ~\citep{tucker1964extension}. We define candidate span-trigger-role triples set  $\mathcal{T}=\{ < s_{i:j}
, r_k, t_e > | s_{i:j} \in \mathcal{S}, r_k \in \mathcal{R}_e \}$. 
Then we introduce a core tensor $\mathbf{Z} \in \mathbb{R}^{d_i\times d^{'}}$ where $d_i$ is the hidden dimension of the combined representation $\mathbf{I}_{i:j}$. 
$\mathbf{I}_{i:j} = FFN([\mathbf{h}_t;\textbf{s}_{i:j}])$, and $FFN$ is the full-connected neural network. For simplity, the
validity for each triple $< s_{i:j}
, r_k, t_e > \in \mathcal{T}$ is scored as follows:

\begin{equation}
    \label{eq17}
    \varphi(s_{i:j}, r_k, t_e)\;=\;\sigma(\mathbf{Z}\;\times_1\;\mathbf{I}_{i:j}\times_{2}\;{\widehat{\mathbf r}}_{\mathrm k}\;+b_k),
\end{equation}
where ${\widehat{\mathbf r}}_{\mathrm k} = \mathbf{H}^{r'}_{k, :}$ represents the transformed role features and $\times_n$ denotes the tensor product operation along the $n^{th}$ dimension. The calculation of $\varphi$ for all triples is performed concurrently via a batched tensor product technique, and the model is trained employing focal loss as described in reference~\citep{8237586}.

\subsubsection{Training and Inference}

Considering the issue of imbalanced role distribution and the prevalence of negative samples among candidate arguments, we utilize focal loss~\citep{8237586} during the training process. Focal loss uses hyperparameters $\alpha$ and $\gamma$ to control the weighting and scaling of the loss function, which allows us to emphasize the importance of positive samples and makes the model focus more on useful information.

\begin{equation}
\label{eq9}
 \begin{array}{c}
\mathcal{L}_c=-\sum_{s_{i:j} \in \mathcal{S}}\alpha{{\lbrack1-P(\varphi(s_{i:j}, r_k, t_e)\;=y_{s_{i:j}}}))\rbrack}^\gamma \vspace{1.3ex} \\
\cdot\log P(\varphi(s_{i:j}, r_k, t_e)\;=y_{s_{i:j}}).
\end{array}
\end{equation}%

Consistent with the training process, we predict and choose the \textit{argmax} role for each candidate span at inference time. If more than one argument span is predicted with the same role, we reserve all spans for this role at inference time.

\subsection{$\text{CARLG}_\text{prompt}$}

In this section, we apply our CARLG framework to prompt-based EAE approaches, where the architecture is shown in Figure~\ref{fig:fig6}. 
 We choose the architecture of PAIE~\citep{ma-etal-2022-prompt} as the foundation to illustrate the prompt-based baseline.
Here, we only provide a detailed introduction to the parts related to our CCA and RLIG and the other parts are omitted.

 \begin{figure*}[tbp]
    \centering
    \includegraphics[width=0.9\linewidth]{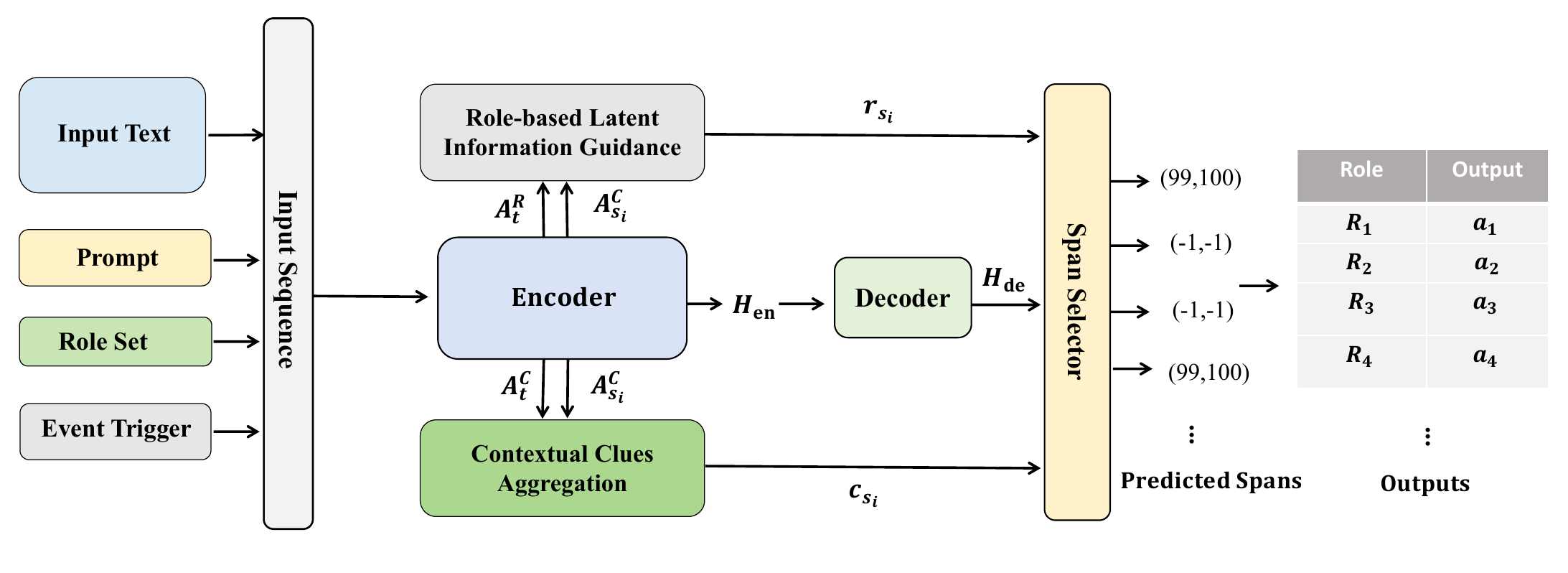}
    
    \caption{The main structure of $\text{CARLG}_\text{prompt}$ involves an input sequence, encompassing roles and prompts, passing through an encoder-decoder architecture. This encoder generates output in the form of context representations, role representations, and attention heads. The Contextual Clue Aggregation (CCA) module dynamically consolidates contextual clues, while the Role-based Latent Information Generation (RLIG) module generates latent role embeddings and captures role correlations. Subsequently, the context vector and the latent role vector are sent to the span selector to predict corresponding spans of each slot.  
Ultimately, based on the predicted spans, we extract the corresponding argument for each role from the text.}
    \label{fig:fig6}
\end{figure*}

\subsubsection{Input Preprocess}

We utilize manual templates to generate the prompt $P$ for each instance\footnote{Inspired by this work~\citep{ma-etal-2022-prompt}, manual template has the most stable performance and usually the better result than other prompt variants.}. The template for each event type is pre-defined as a sentence with slots $\left< role \right> $. 
The generated prompt for each specific event type is a filled template where
slots are substituted with actual role names. We use $s_k$ to denote the $k$-th slot in the prompt $P$. 
\[P = \left< killer \right>  \ killed \ \left< victim \right>  \\ 
using \ instrument \ 
at \left< place \right>.\]
The above case shows a concrete example for the prompt creation, and $s_0$ = $\left< victim \right> $, $s_1$ = $\left< place \right> $, $s_2$ = $\left< killer \right> $.

Next, we concatenate the input text with the prompt. Following Algorithm~\ref{Algorithm 2}, we prepend the role type to the sequence to obtain the final sequence $S_\text{prompt}$.

\subsubsection{Prompt-based CCA and RLIG}

We adopt an encoder-decoder architecture. The encoder is employed to encode the input text, while the decoder is tasked with deriving the event-oriented context and context-oriented prompt
representation $\mathbf{H}_\mathrm{de}$:

\begin{equation}
    \begin{array}{c}
[\mathbf{A}; \mathbf{H}_\mathrm{en}] = \mathrm{Encoder}(S_\text{prompt}), \vspace{1.0ex} 
 \end{array}
\end{equation}
where the $\mathrm{Encoder}$ is a pre-trained transformer-based encoder.
$\mathbf{A} \in \mathbb{R}^{H \times l_s \times l_s}$ is the multi-head attention matrix and $\mathbf{H}_\mathrm{en}$. $H$ is the attention head numbers and $l_s$ is the length of input sequence $S_\text{prompt}$.

For prompt-based EAE approaches, the representation of event arguments is indicated through slots in the prompt. Therefore,  we replace the $\textbf{A}^{C}_{a_i}$ in Algorithm~\ref{Algorithm 1} and $\textbf{A}^{R}_{a_i}$ in Algorithm~\ref{Algorithm 2} with the corresponding attention heads of the slots in the input sequence, respectively.  
The remaining operations are consistent with the steps in Algorithm~\ref{Algorithm 1}  and Algorithm~\ref{Algorithm 2}. Ultimately, we can obtain the enhanced context
vector $\textbf{c}_{s_i}$ and role vector $\textbf{r}_{s_i}$ for the the slot of argument $a_i$.

Then we utilize a decoder to derive the event-oriented context and context-oriented prompt
representation $\mathbf{H}_\mathrm{de}$:

\begin{equation}
    \begin{array}{c}
\mathbf{H}_\mathrm{de} = \mathrm{Decoder} (\mathbf{H}_\mathrm{en}), \vspace{1.0ex} 
 \end{array}
\end{equation}
where the $\mathrm{Decoder}$ is a transformer-based decoder and $\mathbf{H}_\mathrm{de}  \in \mathbb{R}^{l_s \times d}$. 
Next, we integrate $\textbf{c}_{s_i}$ and $\textbf{r}_{s_i}$ into the decoder output $\mathbf{h}_{s{i}} \in \mathbb{R}^{d}$ for slot $s_{i}$, resulting in an updated representation $\tilde{\mathbf{h}}{s{i}} \in \mathbb{R}^{d}$::

\begin{equation} 
\tilde{\mathbf{h}}_{s_{i}}=\mathrm{tanh}(\mathbf{W}^{Prompt}[\mathbf{h}_{s_{i}}; \textbf{c}_{s_i}; \textbf{r}_{s_i} ]),
\end{equation}
where $\mathbf{W}^{Prompt} \in \mathbb{R}^{3d \times d}$ is learnable parameter.

\subsubsection{Span Selection}

Upon acquiring the final representation $\tilde{\mathbf{h}}_{s{i}}$ for every slot across each event, we employ a methodology inspired by~\citep{ma2022prompt} to convert these into a duo of span selectors {$\Phi_{s_{i}}^\mathrm{start}, \Phi_{s_{i}}^\mathrm{end}$}:

\begin{equation}
\begin{array}{c}
\Phi_{s_{i}}^\mathrm{start} = \tilde{\mathbf{h}}_{s{i}} \circ \mathbf{w}_\mathrm{start}, \vspace{1.0ex} \
\Phi_{s_{i}}^\mathrm{end} = \tilde{\mathbf{h}}_{s{i}} \circ \mathbf{w}_\mathrm{end},
\end{array}
\end{equation}
where $\circ$ denotes element-wise multiplication and $\mathbf{w}_\mathrm{start}, \mathbf{w}_\mathrm{end} \in \mathbb{R}^{d}$ are tunable parameters. These selectors, $\Phi_{s_{i}}^\mathrm{start}$ and $\Phi_{s_{i}}^\mathrm{end}$, are pivotal in identifying the start and end points of the slot $s_{i}$ within the original text.
Following the approach in~\citep{ma2022prompt}, the Bipartite Matching Loss is utilized~\citep{carion2020end} to refine the alignment of predicted and actual argument spans during training (
For detailed content on this part, please refer to the work PAIE~\citep{ma-etal-2022-prompt}, and we follow their approach here.)

\section{Experiments}



\subsection{Experimental Setup}
\subsubsection{Datasets}
We evaluate the proposed method on three large-scale document-level EAE datasets, that is, RAMS~\citep{ebner2020multi}, WikiEvents~\citep{lietal2021document} and MLEE~\citep{10.1093/bioinformatics/bts407}, and detailed data statistics are listed in Table~\ref{tab1}.

\begin{table}[]
\caption{Detailed data statistics.}
\centering
\setlength{\tabcolsep}{3mm}{
\begin{tabular}{l|ccc}
\hline
\textbf{Dataset} & \multicolumn{1}{l}{\textbf{RAMS}} & \multicolumn{1}{l}{\textbf{WikiEvents}} & \multicolumn{1}{l}{\textbf{MLEE}} \\ \hline
\textbf{\# Event types}     & 139  & 50   & 23   \\
\textbf{\# Args per event}  & 2.33 & 1.40 & 1.29 \\
\textbf{\# Events per text} & 1.25 & 1.78 & 3.32 \\
\textbf{\# Role types}      & 65   & 57   & 8    \\ \hline
\textbf{\# Events}          &      &      &      \\
\quad Train                       & 7329 & 3241 & 4442 \\
\quad Dev                         & 924  & 345  & -    \\
\quad Test                        & 871  & 365  & 2200 \\ \hline
\end{tabular}
}

\label{tab1}
\end{table}

\begin{itemize} 
\item \textbf{RAMS} \quad The dataset includes 9,124 annotated events sourced from English online news articles. In this dataset, each event forms a single instance, and events that occur within the same context are grouped together into a single instance containing multiple events.  We adopt the original train/dev/test split following~\citep{ebner2020multi}. 
\item \textbf{WikiEvents} \quad
This dataset centers on events documented in English Wikipedia, encompassing both the events and the related news articles referencing these events. It's annotated with co-reference links for arguments. However, for our experiments, we only employ the precise argument annotations provided.
\item \textbf{MLEE} \quad 
        This dataset is a document-level event extraction dataset that focuses on bio-medical publications, consisting of manually annotated abstracts written in English. We follow the procedure outlined in~\citep{10.1093/bioinformatics/btaa540} and it's worth noting that the preprocessed dataset only provides a train/test data split, without a dedicated development set. In this case, we use the training set as the development set for our experiments following~~\citep{10.1093/bioinformatics/btaa540}. 
\end{itemize} 
\subsubsection{Evaluation Metrics}
For the RAMS dataset, we adopt the method detailed in~\citep{xu2022two}, focusing on both Span F1 and Head F1 scores across the development and test sets. Span F1 evaluates the precision of matching predicted argument spans against the gold standard, while Head F1 quantifies the accuracy of identifying the central word in each argument span. For a comprehensive assessment, we also incorporate Span F1 for argument identification (Arg IF) on the test set, as suggested by~\citep{ma-etal-2022-prompt}. Regarding the Wikievents and MLEE datasets, we report Span F1 scores on the test set for both Argument Identification (Arg IF) and Argument Classification (Arg CF) tasks, following the methodologies in~\citep{lietal2021document, ma-etal-2022-prompt}.

\subsubsection{Baselines}
We evaluate various types of document-level Event Argument Extraction (EAE) baselines, including
four span-based methods like \textbf{Two-Step}, $\textbf{Two-Step}_{\text{TCD}}$~\citep{ebner2020multi}, \textbf{TSAR}~\citep{xu2022two}, \textbf{SCPRG}~\citep{liu-etal-2023-enhancing-document}, six generation-based methods such as \textbf{FEAE}~\citep{weietal2021trigger}, \textbf{BART-Gen}~\citep{lietal2021document},
$\textbf{EA}^2$\textbf{E}~\citep{zengetal2022ea2e}, , \textbf{EEQA}~\citep{du-cardie-2020-event}, \textbf{EACE}~\citep{zhou2024eace},  and five prompt-based approaches:  \textbf{PAIE}, \textbf{RKDE}~\citep{hu2023role}, \textbf{HRA}~\citep{ren2023retrieve}, \textbf{SPEAE}~\citep{nguyen2023contextualized},  and \textbf{TableEAE}~\citep{he2023revisiting}. 
To facilitate a straightforward comparison, we utilize the large version of pre-trained language models across various methods for our evaluation. We equip the CARLG framework on the span-based method \textbf{Two-Step} and \textbf{TSAR}, as well as  the prompt-based methods \textbf{PAIE} and \textbf{TabEAE}, and subsequently compare the enhancement in performance with the original methods.

\begin{table}[]
\caption{Hyperparameter settings for $\text{CARLG}_\text{span}$. * means that we tuned the hyperparameters in our experiments. The rest of hyperparameters are set the same as TSAR~\citep{xu2022two}.}
\centering
\setlength{\tabcolsep}{2.4mm}{
\begin{tabular}{cccc}
\hline
\multicolumn{1}{l}{\textbf{Hyperparameters}} & \multicolumn{1}{l}{\textbf{RAMS}} & \multicolumn{1}{l}{\textbf{WikiEvents}} & \multicolumn{1}{l}{\textbf{MLEE}} \\ \hline
Training Epochs    & 50   & 100  & 100  \\
Warmup Ratio       & 0.2  & 0.05 & 0.1  \\
Transformer LR     & 3e-5 & 3e-5 & 3e-5 \\
Not Transformer LR & 1e-4 & 1e-4 & 1e-4 \\
Dropout Ratio      & 0.1  & 0.1  & 0.1  \\
Batch Size         & 4    & 4    & 4    \\
Max Span Length    & 8    & 8    & 8    \\
Max Input Length   & 1024 & 1024 & 1024 \\
Lambda             & 0.05 & 0.1  & 0.05 \\
Alpha*              & 10   & 10   & 10   \\
Gamma*              & 2    & 2    & 2    \\ \hline
\end{tabular}
}

\label{tab:table11}
\end{table} 

\begin{table}[h]
\centering
\setlength{\tabcolsep}{1.5mm}{
\begin{tabular}{lccc}
\hline
\textbf{Hyperparameters}  & \textbf{RAMS} & \textbf{WikiEvents} & \textbf{MLEE} \\ \hline
Training Steps                                          & 10000         & 10000              & 10000         \\
Warmup Ratio*                                               & 0.1           & 0.1                & 0.2           \\
Learning Rate*                 & 2e-5          & 3e-5               & 3e-5          \\
Max Gradient Norm                & 5             & 5                  & 5             \\
Batch Size*                         & 4             & 4                  & 4             \\
Context Window Size              & 250           & 250                & 250           \\
Max Span Length                  & 10            & 10                 & 10            \\
Max Encoder Seq Length           & 500           & 500                & 500           \\
Max Prompt Length*                & 210          & 360                & 360          \\
Encoder Layers*                & 17          & 17                & 17 \\
Decoder Layers*                & 7          & 7                & 7
\\ \hline
\end{tabular}}
\caption{Hyperparameter settings for $\text{CARLG}_\text{prompt}$. * means that we tuned the hyperparameters in our experiments. The rest of hyperparameters are set the same as PAIE~\citep{ma2022prompt}. }
\label{table:hyper}
\end{table}

\subsubsection{Hyperparameter Setting}
Due to the fact that our proposed module does not rely heavily on hyperparameters, most of the hyperparameters follow the same configuration of baselines.
Here, we provide the hyperparameter settings for the span-based method TSAR in Table~\ref{tab:table11} and the prompt-based method PAIE in Table~\ref{table:hyper}

\begin{table*}[]
\caption{The main results for the RAMS dataset. Arg IF denotes the  F1 score for the Argument Identification subtask, and other metrics are for the Argument Classification subtask. \textbf{Bold} indicates the best experimental results. The values in parentheses represent the performance compared to the original methods after incorporating our approach.}
\centering

\setlength{\tabcolsep}{3.5mm}{
\begin{tabular}{lccccc}

\hline
\multirow{2}{*}{\textbf{Method}}  &
  \multicolumn{2}{c}{\textbf{Dev}} &
  \multicolumn{3}{c}{\textbf{Test}} \\ \cline{2-6} 
 &
  \multicolumn{1}{c}{Span F1} &
  \multicolumn{1}{c}{Head F1} &
  \multicolumn{1}{c}{Span F1} &
  \multicolumn{1}{c}{Head F1} &
  \multicolumn{1}{c}{Arg IF} \\ \hline

\textit{Generation-based Methods} \\
${\text{EEQA}}$~\citep{du-cardie-2020-event}                & -              & -              & 46.7     & -     & 48.7              \\
FEAE~\citep{weietal2021trigger}              & -              & -              & 47.4          & -   & -           \\
${\text{BART-Gen}}$~\citep{lietal2021document}    & -              & -              & 48.6          & 57.3      & 51.2    \\

${\text{EA}^2\text{E}}$~\citep{zengetal2022ea2e} &  46.1    & 53.9    & 47.8  & 56.4   &  53.2                            \\

${\text{EACE}}$~\citep{zhou2024eace}                & -              & -              & 50.1          & 56.3     & -  \\

\hline
\textit{Span-based Methods} \\
Two-Step~\citep{zhang2020two}     & 38.9           & 46.4           & 40.1           & 47.7     & 46.5      \\

${\text{Two-Step}}_{{\text{TCD}}}$~\citep{zhang2020two} & 40.3           & 48.0           & 41.8           & 49.7        & -   \\

${\text{TSAR}}$~\citep{xu2022two}        & 49.2          & 56.8          & 51.2          & 58.5     & 56.1     \\

${\text{SCPRG}}$~\citep{liu-etal-2023-enhancing-document}        & 50.2          & 57.2         & 51.9          & 59.1     & -     \\
\hline
\textit{Prompt-based Methods} \\
${\text{PAIE}}$~\citep{ma-etal-2022-prompt}                & 50.9              & 56.5              & 52.2          & 58.8      & 56.8       \\
${\text{RKDE}}$~\citep{hu2023role}                & -              & -              & 50.3          & 55.1     & - \\
${\text{HRA}}$~\citep{ren2023retrieve}                & -              & -              & 48.4     & -     & 54.6              \\
${\text{SPEAE}}$~\citep{nguyen2023contextualized}                & -              & -              & 52.5          & -     & 57.0 \\
${\text{TabEAE}}$~\citep{he2023revisiting}                & 51.9              & 57.8              & 51.8          &  59.0    & 56.7 \\ \hline
\textit{$\text{CARLG}_\text{span}$} (Ours) \\
${\text{Two-Step}}$+$\text{CARLG}_\text{span}$ & 43.4 \footnotesize{($\uparrow$4.5)}           & 50.6 \footnotesize{($\uparrow$4.2)}            & 45.3 \footnotesize{($\uparrow$5.2)}           & 52.8   \footnotesize{($\uparrow$5.1)}      & 50.3 \footnotesize{($\uparrow$3.8)}     \\

${\text{TSAR}}_{{\text{large}}}$+$\text{CARLG}_\text{span}$ & 50.7 \footnotesize{($\uparrow$1.5)}          & {58.1} \footnotesize{($\uparrow$1.3)}          & 52.6 \footnotesize{($\uparrow$1.4)}         &  59.9 \footnotesize{($\uparrow$1.4)}       & {57.1} \footnotesize{($\uparrow$1.0)}   \\
 \hline
\textit{$\text{CARLG}_\text{prompt}$} (Ours) \\
${\text{PAIE}}$+$\text{CARLG}_\text{prompt}$ & 52.8 \footnotesize{($\uparrow$1.9)}           & 58.7 \footnotesize{($\uparrow$2.2)}            & \textbf{54.4} \footnotesize{($\uparrow$2.2)}           & \textbf{60.7}   \footnotesize{($\uparrow$1.9)}      & \textbf{59.1}  \footnotesize{($\uparrow$2.3)}  \\

${\text{TabEAE}}$+$\text{CARLG}_\text{prompt}$ & \textbf{53.1} \footnotesize{($\uparrow$1.2)}          & \textbf{59.2} \footnotesize{($\uparrow$1.4)}          & 54.3 \footnotesize{($\uparrow$2.5)}           & 60.3 \footnotesize{($\uparrow$1.3)}       & 58.6 \footnotesize{($\uparrow$1.9)} \\
\hline
\end{tabular}

\label{tab:table2}
}

\end{table*}

\subsection{Main Results}
\label{exp:main}
In this section, we compare the effectiveness of our CARLG framework on three datasets. We validate the CARLG framework both on span-based and prompt-based methods and then conduct further analyses.
\subsubsection{Performance of Span-based Methods equipped with CARLG}
 As shown in Table~\ref{tab:table2} and Table~\ref{tab:table3}, both Two-Step and TSAR exhibit improved performance when equipped with $\text{CARLG}_\text{span}$. Two-Step yields an improvement of \textbf{+5.2/+5.1} Span F1/Head F1 for Arg CF and \textbf{+3.8} Span F1 for Arg IF on RAMS, \textbf{+6.9/+6.1} Span F1 for Arg IF/Arg CF on WikiEvents, and \textbf{+2.2/+1.9} Span F1 for Arg IF/Arg CF on MLEE.  ${\text{TSAR}}$ achieves notable improvements in Span F1/Head F1 for Arg CF on RAMS (\textbf{+1.4/+1.4}), Span F1 for Arg IF/Arg CF on WikiEvents (\textbf{+1.5/+1.5}), and Span F1 for Arg IF/Arg CF on MLEE (\textbf{+1.3/+1.2}). It is worth noting that the improvement of TSAR after incorporating CCA and RLIG modules is relatively modest. This could be attributed to the fact that the AMR graph already contains contextual semantic information, which may mitigate the need for additional contextual clue integration. These results illustrate that our two modules can be easily applied to other span-based methods with transformer encoders and provide a significant performance boost in document-level EAE tasks, which are compact and transplantable.
\subsubsection{Performance of Prompt-based Methods equipped with CARLG}
When equipped with $\text{CARLG}_\text{prompt}$, as illustrated in Table~\ref{tab:table2} and Table~\ref{tab:table3}, the prompt-based methods PAIE and TabEAE show significant enhancements across all three datasets, surpassing previous generation-based and span-based models. Specifically, PAIE shows a boost of \textbf{+1.1/+1.4} in Span F1/Head F1 for Arg CF and \textbf{+2.3} in Span F1 for Arg IF on RAMS, \textbf{+2.1/+1.7} in Span F1 for Arg IF/Arg CF on WikiEvents, and \textbf{+2.0/+1.3} in Span F1 for Arg IF/Arg CF on MLEE. Meanwhile, ${\text{TabEAE}}$ achieves remarkable improvement in Span F1/Head F1 for Arg CF on RAMS (\textbf{+2.5/+1.3}), in Span F1 for Arg IF/Arg CF on WikiEvents (\textbf{+0.7/+1.1}), and in Span F1 for Arg IF/Arg CF on MLEE (\textbf{+1.1/+0.9}). These results validate the effective adaptability of our CARLG framework to prompt-based EAE methods, showcasing enhanced extraction capabilities.  This is attributed to its adept utilization of contextual clues and the integration of latent role representations enriched with semantic connections.

\begin{table*}[]
\caption{Main results of WikiEvents and MLEE datasets. \textbf{Bold} indicates the best experimental results. The values in parentheses represent the performance compared to the original methods after incorporating our approach.}
\centering

\setlength{\tabcolsep}{4mm}{
\begin{tabular}{lcccc}
\hline
& \multicolumn{2}{c}{\textbf{WikiEvents}} & \multicolumn{2}{c}{\textbf{MLEE}} \\ \cline{2-5} 
\textbf{Method} &
  \textbf{Arg IF} &
  \textbf{Arg CF} &
  \textbf{Arg IF} &
  \textbf{Arg CF} \\ \cline{2-5} 
 &
  Span F1 &
  Span F1 &
  \multicolumn{1}{c}{Span F1} &
  \multicolumn{1}{c}{Span F1} \\ 
\hline
\textit{Generation-based Methods} \\
${\text{EEQA}}$~\citep{du-cardie-2020-event}                & 56.9              & 54.5              & 70.3     & 68.7                  \\

${\text{EA}^2\text{E}}$~\citep{zengetal2022ea2e} & 68.8    & 63.7   & 70.6  & 69.1                              \\

${\text{EACE}}$~\citep{zhou2024eace}                & 71.1              & 66.2              & -          & -     \\

\hline
\textit{Span-based Methods} \\
Two-Step~\citep{zhang2020two}     & 52.9           & 51.1           & 66.7           & 61.9          \\

${\text{Two-Step}}_{{\text{TCD}}}$~\citep{zhang2020two} & 54.1           & 53.1           & 67.2           & 63.0        \\

${\text{TSAR}}$~\citep{xu2022two}        & 70.8          & 65.5         & 72.3          & 71.3         \\

${\text{SCPRG}}$~\citep{liu-etal-2023-enhancing-document}        & 71.3          & 66.4         & -          & -          \\
\hline
\textit{Prompt-based Methods} \\
${\text{PAIE}}$~\citep{ma-etal-2022-prompt}                & 70.5              & 65.3              & 72.1          & 70.8            \\
${\text{RKDE}}$~\citep{hu2023role}                & 69.1             & 63.8             & -          & -     \\
${\text{HRA}}$~\citep{ren2023retrieve}                & 69.6              & 63.4              & - & - \\
${\text{SPEAE}}$~\citep{nguyen2023contextualized}                & 71.9              & 66.1              & -          & -    \\
${\text{TabEAE}}$~\citep{he2023revisiting}                & 71.1              & 66.0              & 75.1          & 74.2     \\ \hline
\textit{$\text{CARLG}_\text{span}$} (Ours) \\
${\text{Two-Step}}$+$\text{CARLG}_\text{span}$ & 59.8 \footnotesize{($\uparrow$6.9)}           & 57.2 \footnotesize{($\uparrow$6.1)}            & 68.9 \footnotesize{($\uparrow$2.2)}           & 63.8   \footnotesize{($\uparrow$1.9)}         \\

${\text{TSAR}}_{{\text{large}}}$+$\text{CARLG}_\text{span}$ & 72.1 \footnotesize{($\uparrow$1.5)}          & {66.8} \footnotesize{($\uparrow$1.5)}          & 73.4 \footnotesize{($\uparrow$1.3)}         &  72.0 \footnotesize{($\uparrow$1.2)}          \\
 \hline
\textit{$\text{CARLG}_\text{prompt}$} (Ours) \\
${\text{PAIE}}$+$\text{CARLG}_\text{prompt}$ & \textbf{72.6} \footnotesize{($\uparrow$2.1)}           & 67.0 \footnotesize{($\uparrow$1.7)}            & 74.1 \footnotesize{($\uparrow$2.0)}           & 73.1   \footnotesize{($\uparrow$1.3)}        \\

${\text{TabEAE}}$+$\text{CARLG}_\text{prompt}$ & 72.1 \footnotesize{($\uparrow$1.0)}          & \textbf{67.1} \footnotesize{($\uparrow$1.1)}          & \textbf{76.2} \footnotesize{($\uparrow$1.1)}           & \textbf{75.1} \footnotesize{($\uparrow$0.9)}       \\
\hline
\end{tabular}
}

\label{tab:table3}
\end{table*}

\subsubsection{Further analyses}
Further analyzing the experimental results in Table~\ref{tab:table2} and Table~\ref{tab:table3}, we observe that: (1) Our CARLG framework shows relatively larger improvement on simple models such as Two-step and PAIE. We hypothesize this may be because other components of complex models  have synergistic or antagonistic effects on  CCA and RLIG modules, constraining the enhancement potential of the CARLG modules and preventing them from fully unleashing their effectiveness. (2) The prompt-based methods yield the best results, with those equipped with the CARLG framework outperforming both generation-based and span-based approaches in terms of performance.


\begin{table*}[]
\caption{Ablation Study on RAMS and WikiEvents dev sets for ${\text{CARLG}}$. 
$\text{PAIE}_\text{C}$ denotes the PAIE method enhanced with our $\text{CARLG}_\text{prompt}$, while $\text{Two-step}_\text{C}$ signifies the Two-step method augmented with our $\text{CARLG}_\text{span}$. Experimental results are averaged based on 5 random seeds.}
\centering
\setlength{\tabcolsep}{3.5mm}{
\begin{tabular}{llcccccccc}
\hline
                                                       &                 & \multicolumn{3}{c}{\textbf{RAMS}}                     & \multicolumn{2}{c}{\textbf{WikiEvents}} \\ \cline{3-7} 
\textbf{Method}                                        & \textbf{Params} & \multicolumn{2}{c}{\textbf{Arg CF}} & \textbf{Arg IF} & \textbf{Arg IF}    & \textbf{Arg CF}    \\ \cline{3-7} 
                                                       &                 & Span F1          & Head F1          & Span F1         & Span F1            & Span F1            \\ \hline
$\text{PAIE}_\text{C}$                                 & 408.22M         & 52.81±0.43       & 58.74±0.36       & 57.03±0.87      & 72.62±0.98         & 67.02±0.85         \\
\quad  \textit{-CCA} &   407.13M         & 51.28±0.60       & 56.67±0.51       & 55.43±0.48      & 70.16±0.87         & 65.87±1.06         \\
\quad  \textit{-RLIG} &  406.14M         & 51.77±0.36       & 56.96±0.44       & 55.91±0.48      & 70.83±0.79         & 66.49±0.68         \\
\quad  \textit{-both} &   405.12M         & 50.94±0.36       & 56.47±0.22       & 55.02±0.24      & 69.85±0.45         & 65.12±0.58         \\ \hline
$\text{Two-step}_\text{C}$                             & 374.70M         & 43.44±0.98       & 50.60±0.77       & 48.78±0.49      & 59.84±1.76         & 57.22±1.26         \\
\quad  \textit{-CCA}   & 373.65M         & 40.24±0.84       & 47.66±0.67       & 46.98±0.26      & 55.48±1.87         & 53.65±1.02         \\
\quad  \textit{-RLIG}                           & 371.68M         & 41.67±1.12       & 48.26±0.95       & 47.24±0.60      & 56.43±1.46         & 54.87±1.19         \\
\quad  \textit{-both} &   370.63M         & 38.93±0.80       & 46.44±0.66       & 46.10±0.90      & 52.92±0.56         & 51.09±1.22         \\ \hline
\end{tabular}
}
\label{tab:table4}
\end{table*}

\subsection{Ablation Study}
To demonstrate the effectiveness of our components, we conduct an ablation study on $\text{CARLG}_\text{span}$ and $\text{CARLG}_\text{prompt}$, utilizing PAIE (denoted as $\text{PAIE}_\text{C}$) for prompt-based methods and Two-step (denoted as $\text{Two-step}_\text{C}$) for span-based methods, with results on the RAMS and WikiEvents datasets presented in Table~\ref{tab:table4}.

Firstly, the removal of the Contextual Clues Aggregation (CCA) module results in a reduction of both Span F1 and Head F1 scores for ${\text{PAIE}}_{{\text{C}}}$ and ${\text{Two-step}}_{{\text{C}}}$ on the RAMS and WikiEvents datasets, respectively. This underscores the pivotal role of our CCA module in capturing vital contextual clues for document-level EAE.

Moreover, when removing the Role-based Latent Information Guidance (RLIG) module\footnote{We additionally remove the respective role tokens that were included in the input sequence.}, we also observe a significant decrease in performance for ${\text{PAIE}}_{{\text{C}}}$ and ${\text{Two-step}}_{{\text{C}}}$. This demonstrates the effect of the RLIG module in capturing meaningful semantic correlations among roles, improving the performance of document-level EAE.
Moreover, when both the CCA and RLIG modules are removed, the decline in performance is even more noticeable in comparison to excluding a single module. This illustrates the synergistic enhancement achieved by the collaboration of these two modules.

    

 \begin{figure*}[tbp]
    \centering
    \includegraphics[width=0.7\linewidth]{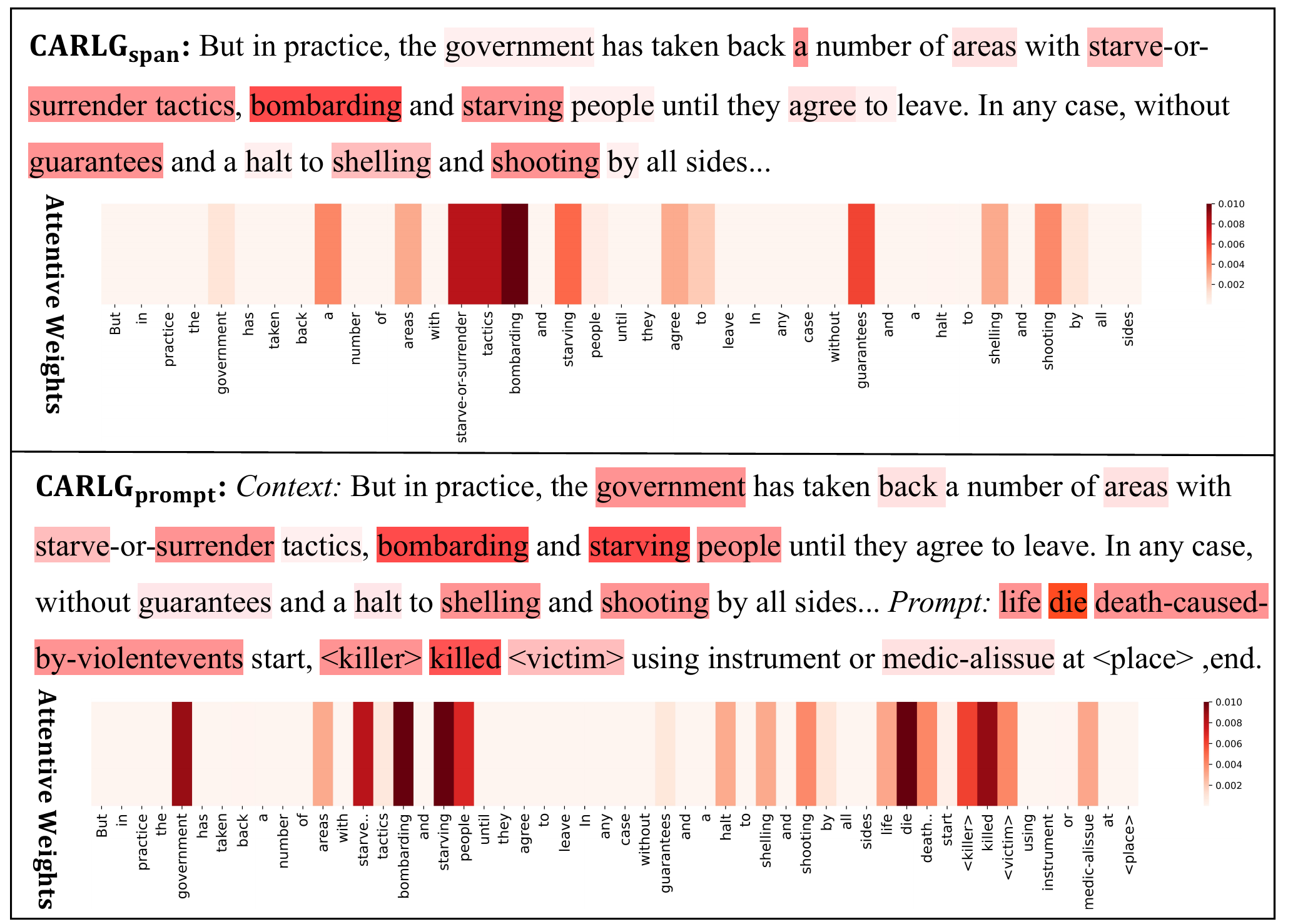}
    
    \caption{Attentive weights for an event instance from RAMS dataset. The event is \texttt{life.die.death-caused-by-violent-events}, focusing on the argument \textit{government} and trigger
\textit{bombarding}). Various color intensities are used to depict the attention weights. We conduct  visualization experiments on $\text{CARLG}_\text{span}$ and $\text{CARLG}_\text{prompt}$, selecting TSAR and PAIE as the backbone methods, respectively.}
    \label{fig:fig6}
\end{figure*}

\subsection{Analysis of Contextual Attentive Weights}
To evaluate the validity of our Contextual Clues Aggregation (CCA) module in capturing beneficial contextual clues for candidate argument spans, we present a visualization of the contextual weights $\textbf{p}^{C}_{a_i}$, as detailed in Eq.\ref{eq4}, using an instance from the RAMS dataset. We conduct  visualization experiments on $\text{CARLG}_\text{span}$ and $\text{CARLG}_\text{prompt}$, selecting TSAR and PAIE as the backbone methods, respectively.
As shown in Figure~\ref{fig:fig6}, the CCA module assigns high weights to non-argument contextual clues such as \textit{surrender}, \textit{starving}, and \textit{shooting}, which are highly relevant to the argument-trigger pair (\textit{government, bombarding}).
Notably, our Contextual Clue Aggregation (CCA) module also grants relatively high attentive weights to some words in other arguments, including \textit{shelling}, \textit{people}. This implies that these argument words offer crucial information for predicting the role of \textit{government}. Additionally, we find that for $\text{CARLG}_\text{prompt}$, the CCA module also learns high attention towards  prompt words such as \textit{killed} and \textit{die}, indicating that the CCA module assists the model in better understanding the prompt. This visualization illustrates how our CCA module effectively identifies relevant contextual clues for target arguments and models the interaction of information among interconnected arguments within an event.

 \begin{figure}[tbp]
    \centering
    \includegraphics[width=1.0\linewidth]{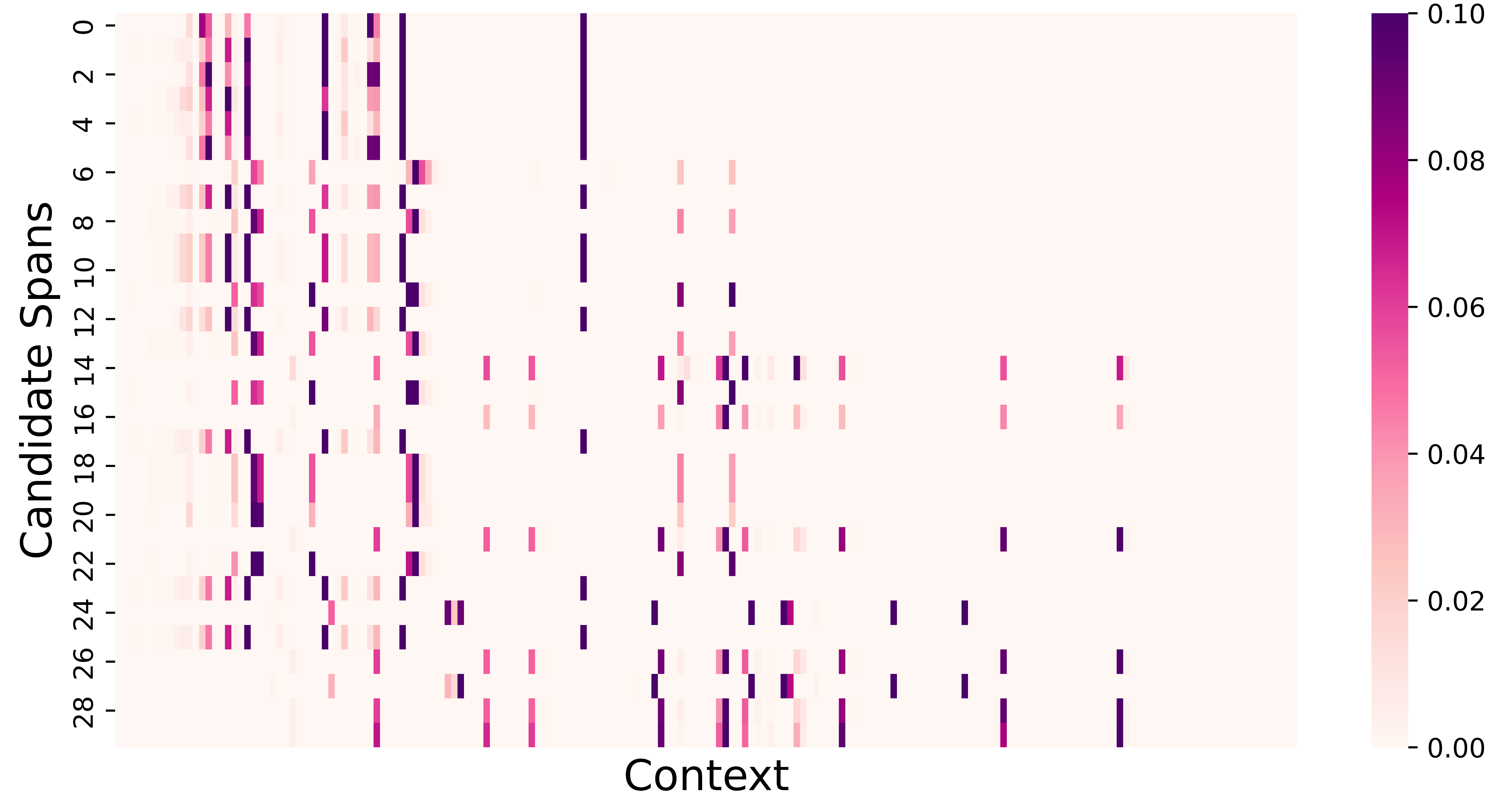}
    
    \caption{We provide visualizations of attention weights directed towards the context for various candidate spans within an event.}
    \label{fig:fig7}
\end{figure}

 Additionally, our investigation extends to examining the attention weights associated with various argument-trigger pairings within a single event.  In Figure~\ref{fig:fig7}, we display a heat map created from the attention weights of 30 randomly chosen candidate spans in an event, with respect to their contextual relevance\footnote{We conduct this experiment on span-based EAE methods TSAR~\citep{xu2022two}  equipped with our $\text{CARLG}_\text{span}$}.  The heat map reveals that different candidate arguments pay attention to different contextual information, suggesting that our Contextual Clues Aggregation (CCA) module is capable of dynamically identifying and utilizing pertinent contextual clues specific to each candidate argument span. This adaptability highlights the module's efficiency in tailoring its focus to suit the unique contextual requirements of each argument span within an event.


\begin{figure}[tbp]
    \centering
    \subfloat[\label{fig:8a } The similarity for $\text{CARLG}_\text{span}$ in RAMS]{
    \includegraphics[width=0.5\linewidth]{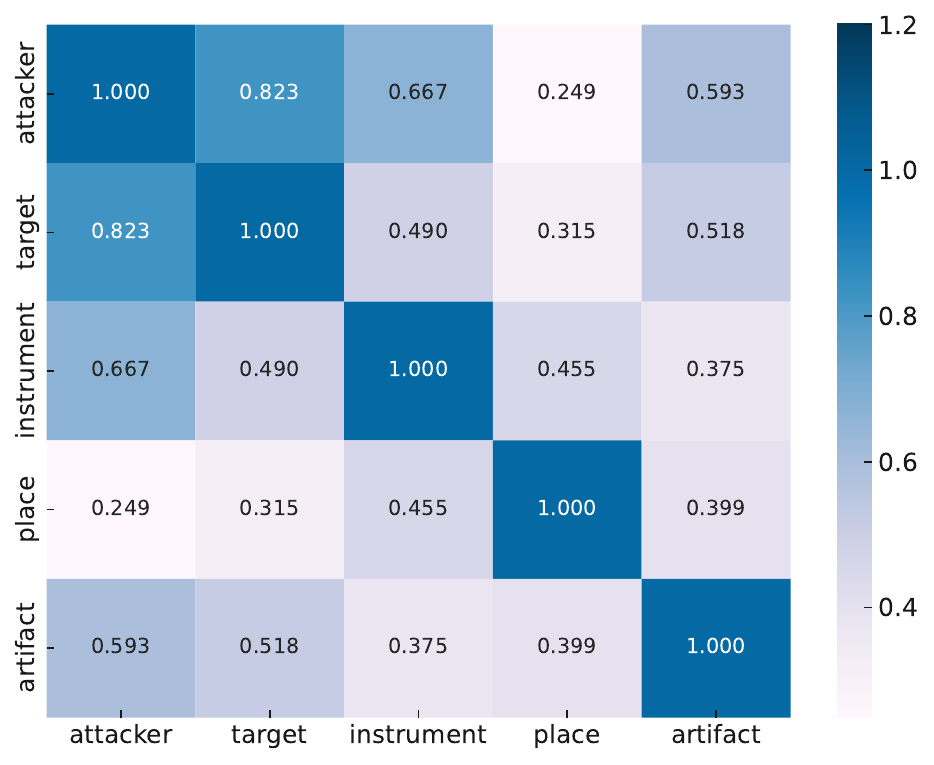}
    }
    \subfloat[\label{fig:8b} The similarity for $\text{CARLG}_\text{prompt}$ in RAMS]{
    \includegraphics[width=0.5\linewidth]{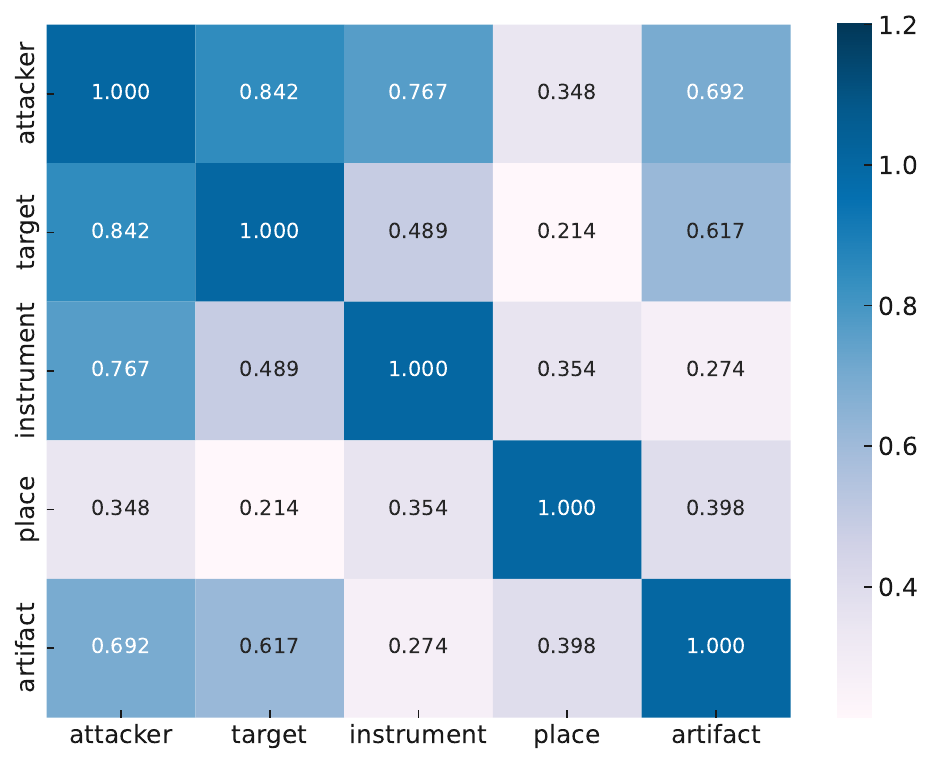}
    }
    \\
     \subfloat[\label{fig:8c} The similarity for $\text{CARLG}_\text{span}$ in MLEE]{
    \includegraphics[width=0.5\linewidth]{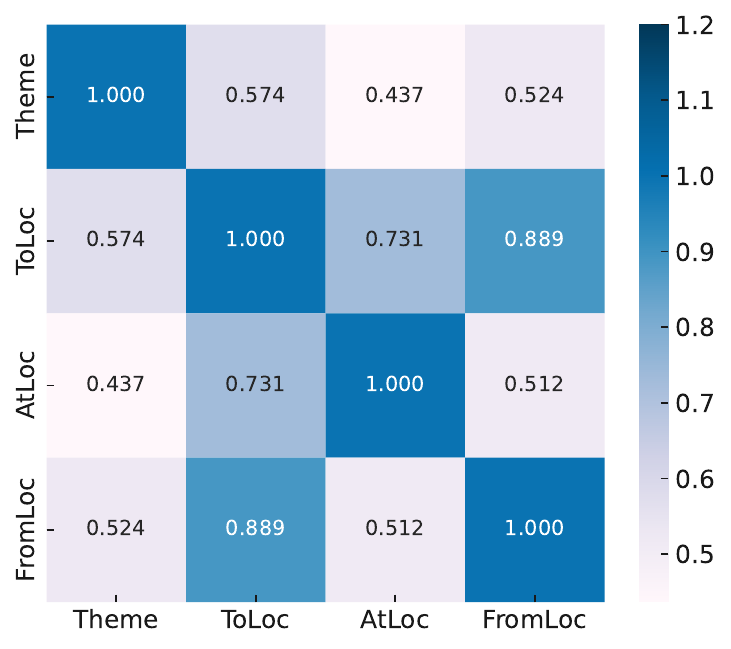}
    }
     \subfloat[\label{fig:8d} The similarity for $\text{CARLG}_\text{prompt}$ in MLEE]{
    \includegraphics[width=0.5\linewidth]{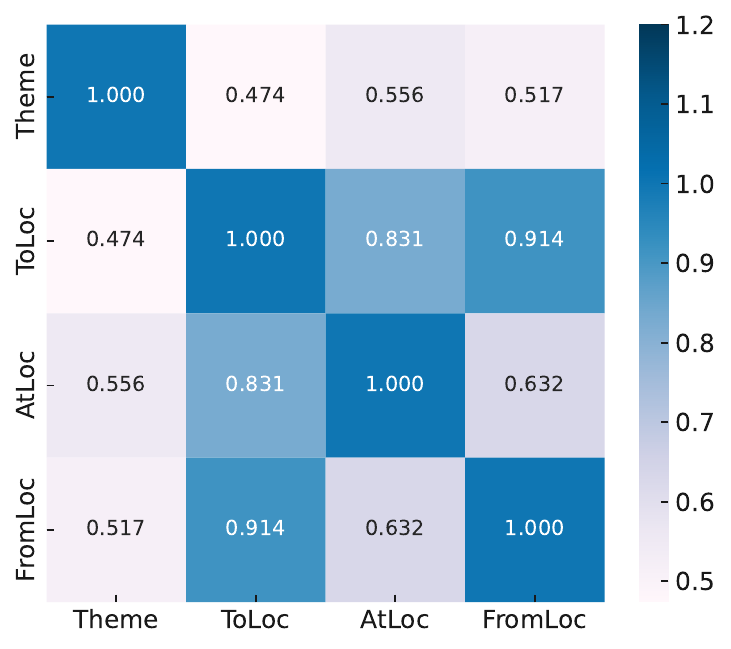}
    }
    \caption{The visualization of cosine similarity between role representations of two examples from RAMS and MLEE datasets, respectively. }
    \label{fig:fig8}
\end{figure}

\begin{figure}[tbp]
    \centering
    \subfloat[\label{fig:9a}]{
    \includegraphics[width=0.8\linewidth]{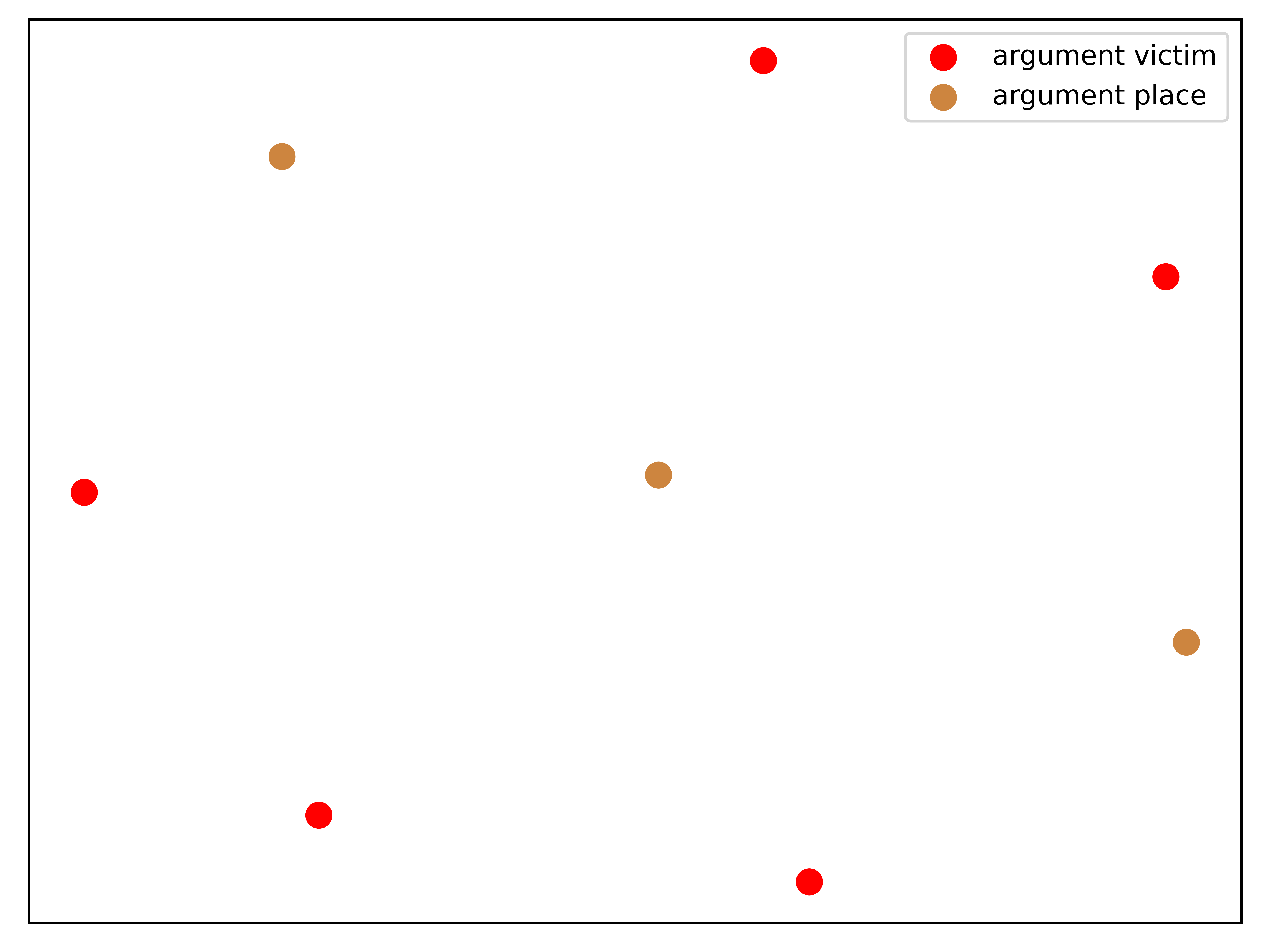}
    } \\
    \subfloat[\label{fig:9b}]{
    \includegraphics[width=0.8\linewidth]{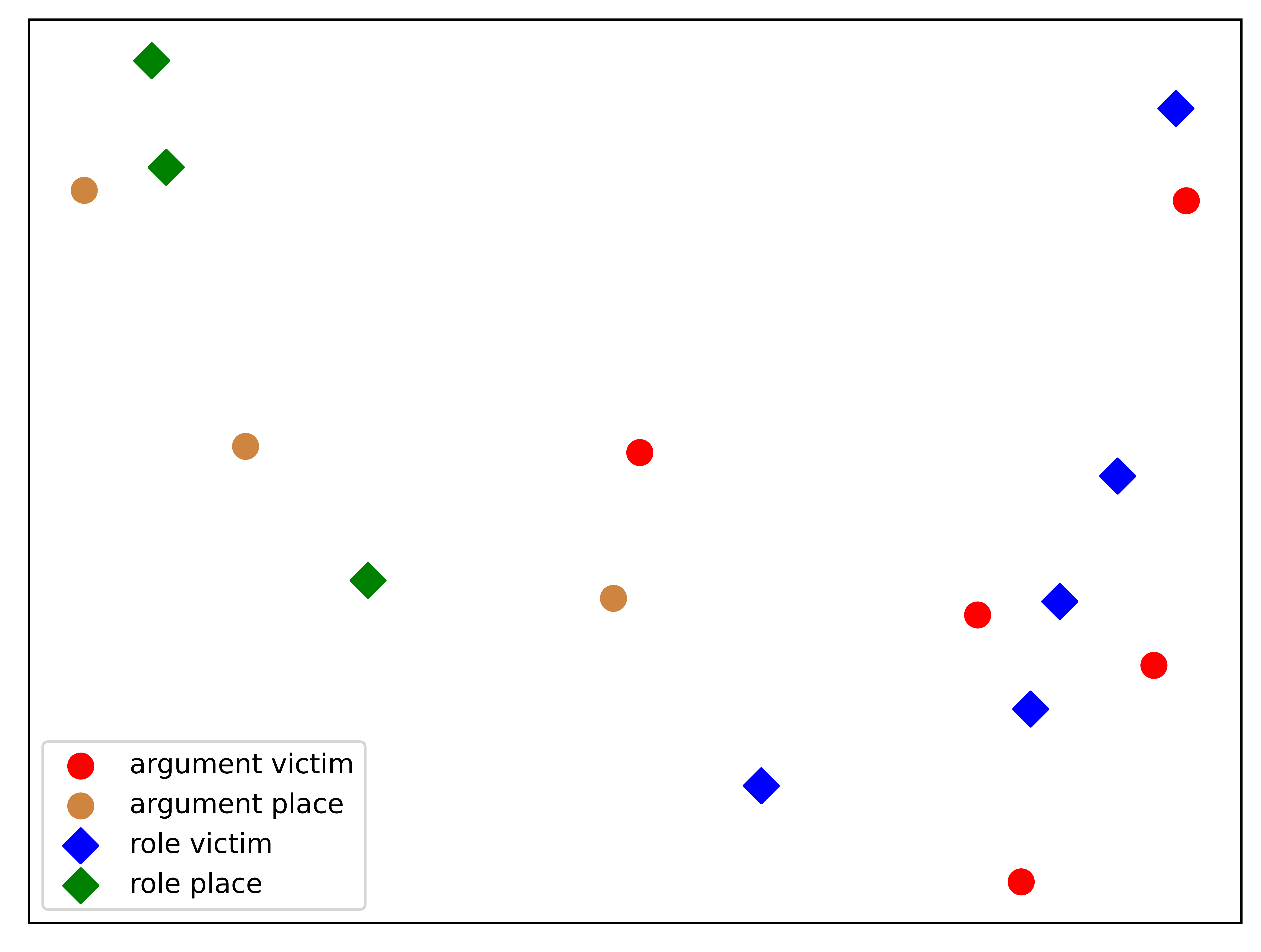}
    }
    \caption{
Here's an instance of a t-SNE visualization sourced from the RAMS dataset. We conduct this experiment using PAIE equipped with our $\text{CARLG}_\text{prompt}$ framawork. The representations of roles and arguments originate from five distinct documents. In sub-figure (a), we utilize average pooling representations encoded by BART~\citep{lewis2019bart}, while in (b), we visualize the representations fused with latent role embeddings.
 }
    \label{fig:fig9}
\end{figure}

\subsection{Analysis of Role Correlations}

To verify the capability of our model in capturing semantic correlations between roles, we visualize the cosine similarity among latent role representations. We select two events within the RAMS and MLEE datasets for $\text{CARLG}_\text{span}$ and $\text{CARLG}_\text{prompt}$\footnote{We select TSAR and PAIE as the backbone methods, respectively.}, respectively. Figure~\ref{fig:fig8} presents the results, where roles such as \textit{origin} and \textit{destination}, \textit{attacker} and \textit{target},  \textit{ToLoc} and \textit{FromLoc} exhibit similar representations, indicating their semantic similarity.
This visualization validates our model's efficacy in capturing semantic correlations among roles.

Additionally, to substantiate the effective guidance provided by role representations, we utilize t-SNE visualization~\citep{JMLR:v9:vandermaaten08a} to display arguments corresponding to two distinct roles that appear together in five diverse documents, alongside their respective latent role embeddings. Figure~\ref{fig:9a} illustrates that arguments associated with the same role but in distinct documents are dispersed throughout the embedding space\footnote{We conduct this experiment using PAIE~\citep{ma-etal-2022-prompt} equipped with our $\text{CARLG}_\text{prompt}$ framawork}. This dispersion is attributed to variations in the target events and their contextual settings. However, after integrating latent role embeddings, the representations of arguments linked to identical roles, such as \textit{victim} or \textit{place}, cluster more closely. This shows that the Role-based Latent Information Guidance (RLIG) module effectively provides crucial role-specific information. The visualization demonstrates that the RLIG module enhances the alignment of arguments within the same role across different contexts, thereby affirming its utility in providing insightful guidance of roles.

\subsection{Complexity and Efficiency}
Our CARLG method combines computational efficiency with low complexity while both CCA and RLIG introduce few parameters.
As shown in Table~\ref{tab:table4}, for  CCA utilizes the attention heads of the pre-trained encoder  and performs multiplication and normalization operations, adding less than 0.5\% new parameters. RLIG introduces approximately 0.8\% new parameters totally. This makes our parameter quantity 
 approximately equivalent to  the transformer-based encoder plus a MLP classification layer.

 \begin{figure}[tbp]
    \centering
    \includegraphics[width=0.9\linewidth]{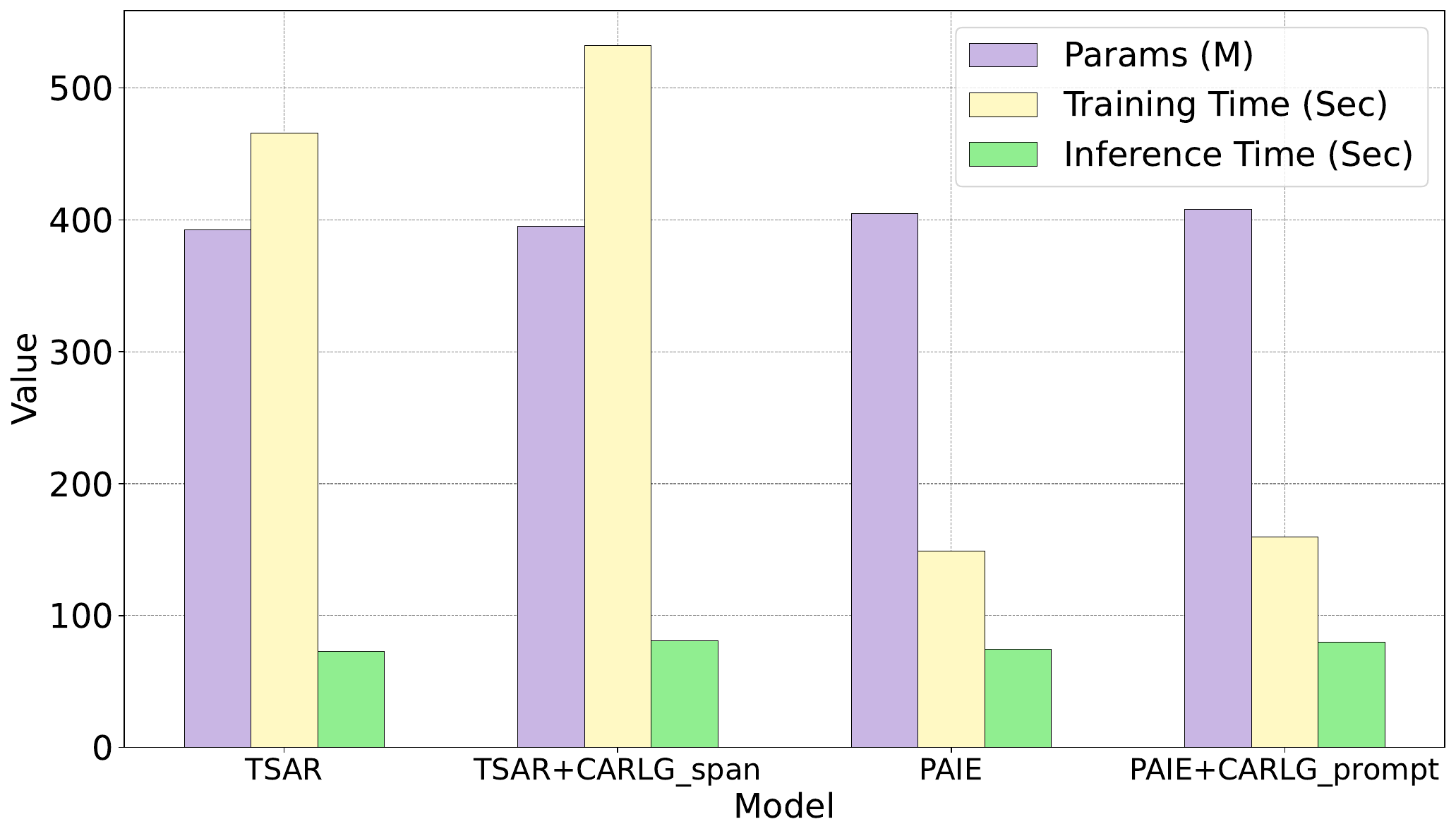}
    
    \caption{The parameter size, training time, and inference time of TSAR and PAIE before and after incorporating CARLG.}
    \label{fig:zhexian}
\end{figure}

Table~\ref{tab:table6} reports the efficiency comparison of various span-based models on MLEE dataset.  After integrating our $\text{CARLG}_\text{span}$ framework, ${\text{TSAR}}_{\text{C}}$ shows an increase of approximately 14.2\% in training time and 11.4\% in inference time. With the addition of $\text{CARLG}_\text{prompt}$,  ${\text{PAIE}}_{\text{C}}$ shows an increase of about 7.4\% in both training and inference time.  
The increased training and inference time associated with $\text{CARLG}_\text{span}$ is attributed to the exhaustive nature of the span-based EAE approaches, necessitating the enumeration of all possible spans. Consequently, the operations of the CCA and RLIG modules escalate in frequency as the quantity of candidate spans expands.

\begin{table}[]
\caption{
The parameter size and time cost (in seconds) before and after integrating the PAIE and TSAR models with the CARLG framework on 
MLEE dataset. Experiments are run for one epoch on one
Tesla A100 (40G) GPU.}
\centering
\small{
\setlength{\tabcolsep}{0.5mm}{
\begin{tabular}{lccc}
\hline
\textbf{Model} & \textbf{Params} & \textbf{Training Time}
& \textbf{Inference Time} \\ \hline
${\text{TSAR}}$   & 392.44M        & 465.84                 & 72.79                   \\

${\text{PAIE}}$   & 405.12M        & 148.74                 & 74.31                  \\


$\text{TSAR}_\text{C}$    & 395.23M \footnotesize{($\uparrow$0.71\%)}      & 532.08  \footnotesize{($\uparrow$14.22\%)}                & 81.08 \footnotesize{($\uparrow$11.39\%)}                   \\ 
$\text{PAIE}_\text{C}$    & 408.22M \footnotesize{($\uparrow$0.77\%)}      & 159.80   \footnotesize{($\uparrow$7.44\%)}              & 79.81  \footnotesize{($\uparrow$7.40\%)}                 \\ 
\hline
\end{tabular}
}
}

\label{tab:table6}
\end{table}

\subsection{Error Analysis}

\begin{table*}[tbp]
\caption{Error analysis on the RAMS test set. ${\text{TSAR}}_{\text{C}}$ and ${\text{PAIE}}_{\text{C}}$ represent the original TSAR and PAIE methods equipped with our CARLG framework. The best results are in bold font.}
\centering
\small{
\setlength{\tabcolsep}{3mm}{
\begin{tabular}{lcccccc}
\hline
\footnotesize{\textbf{Model}} & \footnotesize{\textbf{Wrong Span}} & \footnotesize{\textbf{Over-extract}} & \footnotesize{\textbf{Partial}} & \footnotesize{\textbf{Overlap}} & \footnotesize{\textbf{Wrong Role}} & \footnotesize{\textbf{Total}} \\ \hline
Two-Step & 86 & 64 & \textbf{47} & 32 & 46 & 275 \\
${\text{TSAR}}$     & 81 & 48 & 57 & \textbf{28} & 19 & 233 \\

${\text{TSAR}}_{\text{C}}$ (Ours)   & \textbf{79} & \textbf{45} & 55 & 29 & \textbf{15} & \textbf{223} \\ \hline
${\text{PAIE}}$     & 61 & 46 & 59 & {34} & 25 & 225 \\
${\text{PAIE}}_{{\text{C}}}$ (Ours)    & \textbf{59} & \textbf{45} & \textbf{59} & \textbf{29} & \textbf{18} & \textbf{210} \\ 
\hline
\end{tabular}
}
}
\label{tab:table7}
\end{table*}

 \begin{figure*}[htbp]
    \centering
    \includegraphics[width=0.8\linewidth]{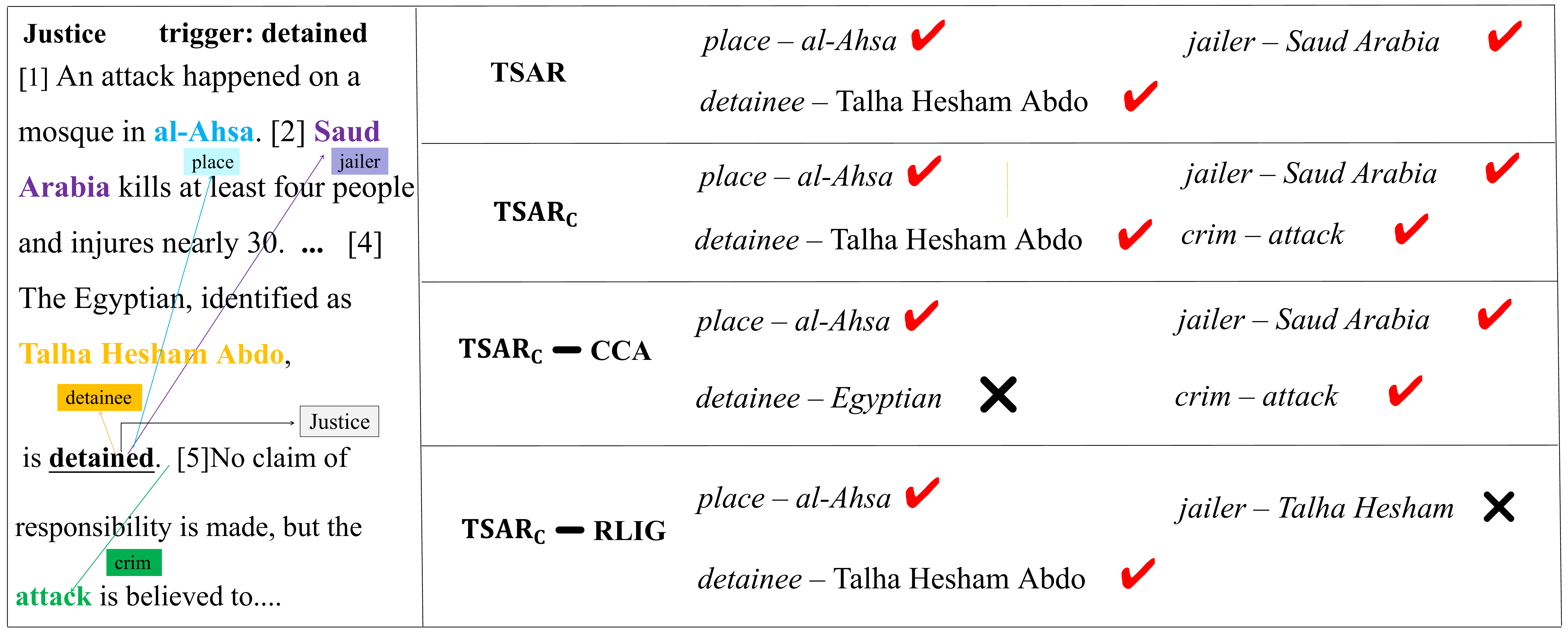}
    
    \caption{An extraction case from RAMS test set base on ${\text{Roberta}}_{{\text{large}}}$. This case shows both CCA module and RLIG module contribute to the argument extraction.}
    \label{fig:fig10}
\end{figure*}
To conduct a more comprehensive examination of the errors made by various models and analyze the underlying reasons for these errors
, we manually compared the predictions of 200 randomly selected examples from the RAMS test set against their golden annotations. During our analysis, we categorized the errors into five main categories following~\citep{xu2022two}. \textbf{Wrong Span} refers to the case where a specific role is assigned to a span that does not overlap with the corresponding golden span. We observe that this type of errors often occur when negative words such as ``not'' are involved or when there are coreference spans related to the golden span. 
The category labeled \textbf{Over-extract} pertains to instances where the model predicts an argument role that does not genuinely appear in the document. Some of the extracted spans are either sub-strings of the golden spans (\textbf{Partial}) or exhibit overlaps with them (\textbf{Overlap}).
 The ``Partial" error category is also commonly observed when there is punctuation, such as a comma, within the golden span. \textbf{Wrong Role} refers to the model correctly identifies the golden span, but assigns the wrong argument role to it. 
 
 We compare the number of errors on TSAR and PAIE before and after adding the CARLG framework.  As shown in Table~\ref{tab:table7}, compared with ${\text{Two-Step}}$ and ${\text{TSAR}}$, our ${\text{TSAR}}_{\text{C}}$ decreases the number of errors from 275 to 223 and effectively reduces Wrong Span errors, which can be attributed to the integration of contextual clues and latent role guidance. Meanwile, compared with the original PAIE method,  the number of errors decreases from 225 to 210 after adding our $\text{CARLG}_\text{prompt}$, especially Wrong Role and Overlap errors. This fully demonstrates that our proposed RLIG module can capture semantic correlations between roles.

\begin{table}[tbp]
\centering
\caption{Comparison with large language model methods on the test set of  RAMS.}
\setlength{\tabcolsep}{4.0mm}{
\begin{tabular}{lll}
\hline
\multirow{2}{*}{Method} & \multicolumn{2}{c}{RAMS} \\
                        & Arg-I       & Arg-C      \\ \hline             
text-davinci-003        & 46.1       & 39.5      \\
gpt-3.5-turbo           & 38.3       & 31.5       \\
gpt-4                   & 50.4        & 42.8      \\ \hline
$\text{TSAR}_\text{C}$ (Ours)                  & 57.1        & 52.6       \\ 
$\text{PAIE}_\text{C}$ (Ours)                  & 59.1        & 54.4  
  \\ \hline
\end{tabular}
}

\label{tab:llm}
\end{table}

\subsection{Comparison with Large Language Models}

Large language models (LLMs) have become a focal point of interest among researchers, given their remarkable versatility across a diverse range of Information Extraction (IE) tasks~\citep{xu2023large, li2023semi, zhang2024ultra}. In this paper, we undertake a comprehensive comparison with some large language models on the document-level EAE task.  Due to the high cost associated with using
GPT-4, its evaluation is limited solely to the RAMS
dataset. The experimental results are detailed in Table~\ref{tab:llm}. We utilize the prompts defined in the work~\citep{zhou2023heuristics}
and  experiments are conducted utilizing three prominent large language models: text-davinci-003~\citep{ouyang2022training}, gpt-3.5-turbo~\citep{ouyang2022training} and GPT-4~\citep{achiam2023gpt}, all of which are accessed via OpenAI’s public APIs~\footnote{\url{https://openai.com/api/}}.
As illustrated in Table~\ref{tab:llm}, LLMs exhibit a notable performance disparity compared to supervised learning models in the EAE task. Furthermore, the operational costs associated with large models are inherently substantial. In contrast, our approach demonstrates superior efficiency, cost-effectiveness, and overall effectiveness in the document-level EAE task when compared to LLMs.

\subsection{Case Study}
In order to explore how the CCA and RLIG modules work in actual cases,  we conduct the case study for our CARLG\footnote{We present a case study on $\text{CARLG}_\text{span}$, choosing TASR~\citep{xu2022two} as the backbone model.}. As shown in Figure~\ref{fig:fig10}, equipped with CCA and RLIG modules, our CARLG can correctly predict all arguments, while TSAR fails to predict arguments
 \textit{attack}. When removing the CCA module, our CARLG fails to predict the argument~\textit{Talha Hesham Abdo}, indicting the CCA module contributes to the argument extraction by providing essential contextual clues. Additionally, when the RLIG module is removed, it fails to predict the argument for the \textit{crim} role and mispredicts the argument for the \textit{jailer} role. This can be attributed to that the RLIG module effectively captures the semantic correlations among the roles \textit{jailer}, \textit{detainee}, and \textit{crim} and provides valuable information guidance, allowing the model to make more accurate predictions. 
\section{Conclusion}
In this paper, we introduce the CARLG model for document-level event argument extraction. CARLG comprises two novel modules: Contextual Clues Aggregation (CCA) and Role-Based Latent Information Guidance (RLIG). CCA module effectively aggregates non-argument clue words to capture beneficial contextual information, while RLIG module provides guidance based on latent role representations that capture semantic correlations between some roles.
Diverse experiments show that our CARLG model surpasses current SOTA EAE models on three large-scale document-level datasets.

While our CARLG model has exhibited remarkable performance in document-level event argument extraction tasks with known event triggers, we acknowledge its limitation in depending on the availability of event triggers. In practical situations, event triggers may not always be supplied or known in advance. Therefore, in our future research, we intent to transfer our approach to encompass other EAE tasks that do not rely on explicit trigger words. We aim to conduct extensive experiments to assess its performance in such scenarios.
Furthermore, we intend to explore the applicability of the CARLG framework in other information extraction tasks, such as relation extraction.


\printcredits

\bibliographystyle{model5-names}

\bibliography{cas-refs}


\end{document}